%% file: main.tex
\documentclass[final]{cvpr}

\usepackage{times}
\usepackage{epsfig}
\usepackage{graphicx}
\usepackage{amsmath}
\usepackage{amssymb}
\usepackage{bm}
\usepackage{bbm}
\usepackage{graphics}
\usepackage{enumitem}
\usepackage[dvipsnames]{xcolor}

\usepackage[skip=0pt]{caption}
\usepackage[pagebackref=true,breaklinks=true,colorlinks,bookmarks=false]{hyperref}

\def\cvprPaperID{1934} 
\def\confYear{CVPR 2021}

\begin{document}

\title{DECOR-GAN: 3D Shape Detailization by Conditional Refinement}

\author{
Zhiqin Chen $^{1}$
\and
Vladimir G. Kim $^{2}$
\and
Matthew Fisher $^{2}$
\and
Noam Aigerman $^{2}$
\and
\hspace{3cm}
\and
Hao Zhang $^{1}$
\and
Siddhartha Chaudhuri $^{2,3}$
\and
\hspace{3cm}
\and
$^{1}$ Simon Fraser University\\
[-3.0ex]
\and
$^{2}$ Adobe Research\\
[-3.0ex]
\and
$^{3}$ IIT Bombay\\
[-3.0ex]
}

\maketitle

\begin{abstract}
\input{0_abstract}
\end{abstract}

\input{1_intro}

\input{2_related}

\input{3_method}

\input{4_results}

\input{5_future}

\input{6_acks}

{\small
\bibliographystyle{ieee_fullname}
\bibliography{mainbib}
}

\clearpage
\input{6_supplementary}

\end{document}

%% file: 0_abstract.tex
We introduce a deep generative network for {\em 3D shape detailization\/}, akin to stylization with the style being geometric details. We address the challenge of creating large varieties of high-resolution and detailed 3D geometry from a small set of exemplars by treating the problem as that of {\em geometric detail transfer\/}. Given a low-resolution coarse voxel shape, our network {\em refines\/} it, via voxel upsampling, into a higher-resolution shape enriched with geometric details. The output shape preserves the overall structure (or content) of the input, while its detail generation is {\em conditioned on\/} an input ``style code'' corresponding to a detailed exemplar.
Our 3D detailization via conditional refinement is realized by a generative adversarial network, coined DECOR-GAN. The network utilizes a 3D CNN generator for upsampling coarse voxels and a 3D PatchGAN discriminator to enforce local patches of the generated model to be similar to those in the training detailed shapes. During testing, a style code is fed into the generator to condition the refinement.
We demonstrate that our method can refine a coarse shape into a variety of detailed shapes with different styles. The generated results are evaluated in terms of content preservation, plausibility, and diversity. Comprehensive ablation studies are conducted to validate our network designs.
Code is available at \href{https://github.com/czq142857/DECOR-GAN}{https://github.com/czq142857/DECOR-GAN}.

%% file: 1_intro.tex
\vspace{-10pt}
\section{Introduction}
\label{sec:intro}

\input{figures/teaser.tex}

Creating high-quality detailed 3D shapes for visual design, AR/VR, gaming, and simulation is a laborious process that requires significant expertise. Recent advances in deep generative neural networks have mainly focused on learning low-dimensional~\cite{PCGAN,BSPNet,AtlasNet,OCNN,wu2016learning} or structural representations~\cite{3DGenSurvey,GRASS,StructureNet,CompME,PQNet} of 3D shapes from large collections of stock models, striving for plausibility and diversity of the generated shapes.
While these techniques are effective at creating coarse geometry and enable the user to model rough objects, they lack the ability to represent, synthesize, and provide control over the finer geometric details. 

In this work, we pose the novel problem of \emph{3D shape detailization}, akin to stylization with the style defined by geometric details. We wish to address the challenge of creating large varieties of high-resolution and detailed 3D geometries from only
a small set of detailed 3D exemplars by treating the problem as that of {\em geometric detail transfer.}
Specifically, given a low-resolution coarse \emph{content} shape and a \emph{detailed style} shape, we would like to synthesize a novel shape that preserves the coarse structure of the content, while refining its geometric details to be similar to that of the style shape; see Figure~\ref{fig:teaser}. Importantly, the detail transfer should {\em not\/} rely on any annotations regarding shape details.

Our conditional detailization task cannot be accomplished by simply copying local patches from the style shape onto the content shape, since (a) it is unknown which patches represent details; (b) it may be difficult to integrate copied patches into the content shape to ensure consistency.
To this end, we train a {\em generative\/} neural network that learns detailization priors over a collection of high-resolution detailed exemplar shapes, enabling it to refine a coarse instance using a \emph{detail style code}; see top of Figure~\ref{fig:teaser}. 

To date, there has been little work on generating high-resolution detailed 3D shapes. For example, surface-based representations that synthesize details on meshes~\cite{Hertz2020deep,Liu20} lack the ability to perform topological changes (as in several plant detailization results in Figure \ref{fig:teaser}) and, due to complexities of mesh analysis, do not consider large shape contexts, limiting their notion of details to homogeneous geometric texture. 
To allow topological variations 
while leveraging a simpler domain for analyzing the content shape, we choose voxel representations for our task. However, 3D grids do not scale well with high resolution, which necessitates a network architecture that effectively leverages limited capacity. Also, regardless of the representation, careful choices of network losses must be made to balance the conflicting goals of detailization and (coarse) content preservation.

To tackle all these challenges, we design DECOR-GAN, a novel generative adversarial network that utilizes a 3D CNN generator to locally
refine a coarse shape, via voxel upsampling, into a high-resolution detailed model, and a 3D PatchGAN discriminator to enforce local patches of the generated model to be similar to those in the training detailed shapes.
Our generator learns local filters with limited receptive field, and thus effectively allocates its capacity towards generating local geometry. We condition our refinement method on a single detail style code that provides an intuitive way to control the model while also ensuring that generated details are consistent across the entire shape.
We train our approach adversarially, using masked discriminators to simultaneously ensure that local details are stylistically plausible and coherent, while the global shape (i.e., downsampled version of the detailized shape) still respects the coarse source shape.
Our method can be trained even with a small number of detailed training examples (up to 64 for all of our experiments), since our convolutional generator only relies on learning local filters. 

We demonstrate that DECOR-GAN can be used to refine coarse shapes derived by downsampling a stock model or neurally generated by prior  techniques. The user can control the style of the shape detailization either by providing a style code from an exemplar or by interpolating between existing styles. Thus, our technique offers a {\em complementary latent space for details} which could be used jointly with existing techniques that provide latent spaces for coarse shapes. We quantitatively evaluate our method using measures for content preservation, plausibility, and diversity. We also provide a comprehensive ablation study to validate our network designs, while demonstrating that simpler approaches (such as only using a reconstructive loss) do not provide the same quality of generated details.

%% file: figures/teaser.tex
\begin{figure}[t!]
\begin{center}
\includegraphics[width=1.0\linewidth]{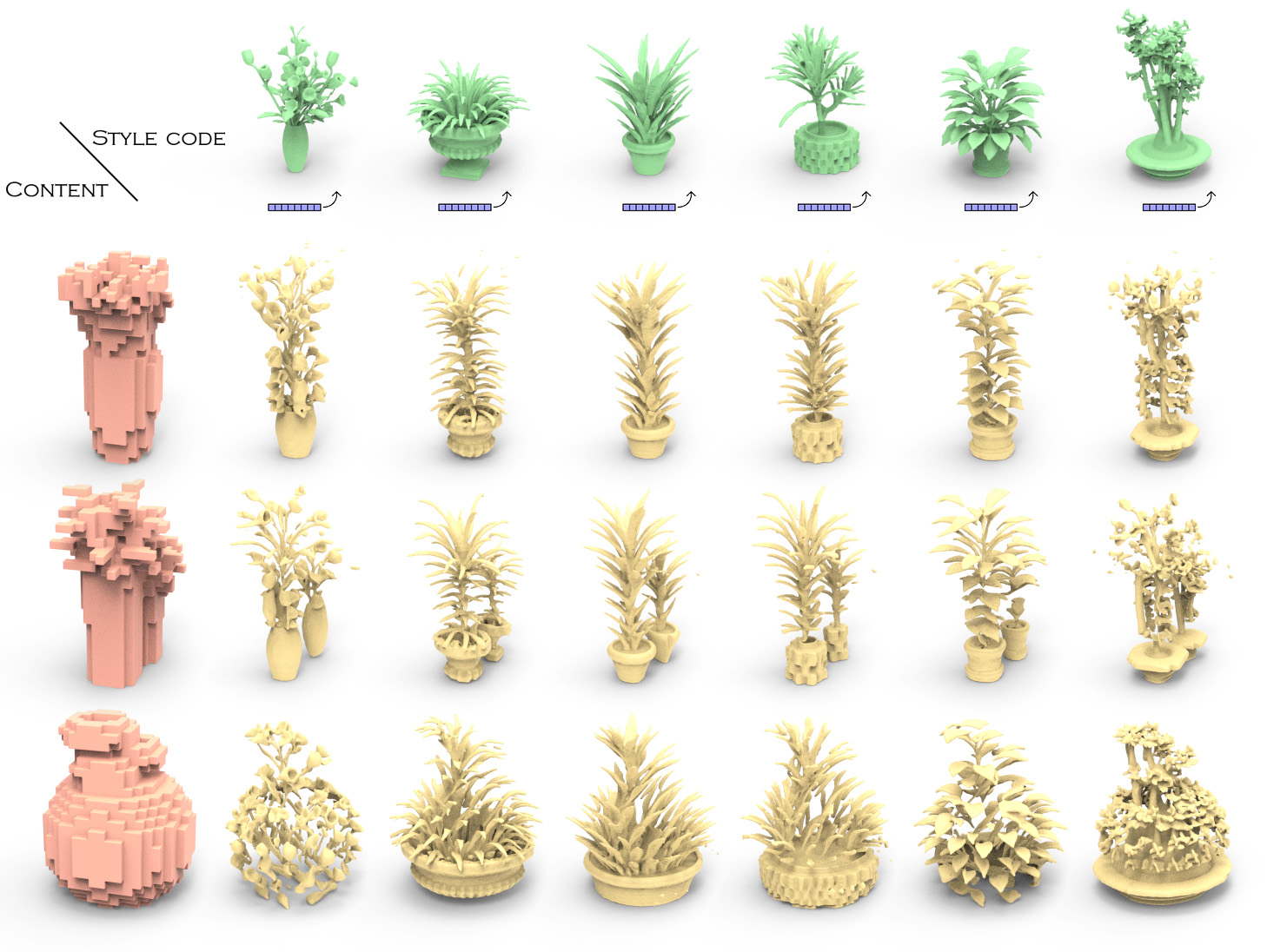}
\end{center}
\caption{
Our 3D detailization network, DECOR-GAN, refines a coarse shape (red, leftmost) into a variety of detailed shapes, each conditioned on a style code characterizing an exemplar detailed 3D shape (green, topmost).
\vspace{-2mm}
}
\label{fig:teaser}
\end{figure}

%% file: 2_related.tex
\section{Related work}
\label{sec:related}

\input{figures/outline_net}

Deep learning models have explored many possible representations for understanding and generating 3D content. These include encoders and decoders for voxel~\cite{brock2016generative,choy20163d,wu2016learning}, point cloud~\cite{fan2017point,qi2017pointnet}, mesh~\cite{gkioxari2019mesh,nash2020polygen,wang2018pixel2mesh}, atlas chart~\cite{AtlasNet,yang2018foldingnet}, and implicit occupancy function representations~\cite{IMNET,mescheder2019occupancy,DISN}.

\vspace{-3mm}

\paragraph{High-Resolution 3D Representations.}
While most representations can approximate the coarse shape of an object, specialized techniques have been developed for each representation to improve the quality of surface geometric detail, and hallucinate high-frequency detail from low-frequency input (``super-resolution''). For implicit functions, detail has been improved by training a coarse model, then increasing the accuracy and focusing on local details~\cite{duan2020curriculum}. An alternative approach uses convolutional layers to generate input-point-specific features for the implicit network to decode~\cite{peng2020convolutional}. Single-view 3D reconstruction quality has also been improved by using local image features to recover geometric detail in the implicit network~\cite{DISN}. For voxel representations, several hierarchical methods increase quality and resolution~\cite{hane2017hierarchical,riegler2017octnet,OCNN}. Smith et al.~\cite{smith2018multiview} take a different approach by reconstructing a voxel grid from high-resolution 2D depth maps. Taking this further, Mildenhall et al.~\cite{mildenhall2020nerf} acquire high-resolution 3D geometry by merging several calibrated images into a neural implicit function. A PatchMatch approach has been used to reconstruct a partial 3D surface scan by directly copying patches from training models~\cite{dai2017shape}. Wang et al.~\cite{yifan2019patch} proposed a patch-based upscaling approach to super-resolve point clouds. Although these methods improve surface detail, they cannot be conditioned on an input style code and do not easily allow for controlled surface detail generation.

Mesh-based learning methods have been proposed to adjust the subdivision process of an input mesh to control surface details~\cite{Liu20}, synthesize surface textures using multi-scale neurally displaced vertices~\cite{Hertz2020deep}, or reconstruct a surface with reoccurring
geometric repetitions from a point cloud~\cite{Point2Mesh}. While able to produce highly detailed surfaces, these methods cannot
alter the topology of the input mesh or mix geometric details from a collection of style shapes.
In a similar vein, our 3D detailization solution also differs from conventional image~\cite{SRGAN} or 3D upsampling~\cite{sanchez2018brain} with the added controllability while requiring much fewer detailed high-resolution shapes during training.

\vspace{-3mm}

\paragraph{Shape Detail Transfer.}
Mesh ``quilting'' is an early non-learning-based method for shape detail transfer; it tiles the surface of a coarse shape with copies of a geometric texture patch~\cite{zhou2006quilting}. Takayama et al.~\cite{takayama2011geobrush} transfer a detailed patch from one shape to another by matching parametrizations. Chen et al.~\cite{chen2012meshmatch} extend 2D PatchMatch to surfaces in 3D. These methods are not automatic and/or require explicit factorization of the shape into content and detail layers. Ma et al.~\cite{ma2014analogy} solve the analogy -- \mbox{``shape : target :: exemplar : ?''} -- by automatically assembling exemplar patches. They require accurate surjective correspondences, local self-similarity and, most importantly, {\em three} input shapes to implicitly define the style-content factorization.

Among learning-based methods, Berkiten et al.~\cite{berkiten2017learning} transfer displacement maps from source to target shapes in one-shot fashion. Wang et al.~\cite{yifan2020neural} propose neural cages to warp a detailed object to the coarse shape of another. Neither of these methods can mix details from multiple shapes or synthesize new topology. Chart-based methods~\cite{ben2018multi,AtlasNet} map a common 2D domain to a collection of shape surfaces, which can be used to transfer details to and between shapes. However, they also cannot synthesize new topology, and can accurately represent only relatively simple base shapes.

\vspace{-3mm}

\paragraph{Image Synthesis.}
Several methods control content and style in 2D image generation~\cite{gatys2016image,li2017universal,park2020swapping}. Recent approaches to generative imaging~\cite{fish2020sketchpatch,Karras2019stylegan2,park2020swapping} employ a PatchGAN discriminator~\cite{isola2017image} that has also been used to construct a latent space for 3D shape categories~\cite{wu2016learning}. One effective way to condition a generative image model is to inject a latent code into each level of the generator~\cite{dumoulin2016learned,park2019gaugan}. We build upon these techniques in our generator design, which constructs a latent space over shape detail that is injected into the generator architecture to guide detail synthesis. We similarly employ a PatchGAN to ensure that the generated 3D shape resembles patches from training shapes.

%% file: figures/outline_net.tex
\begin{figure*}[t!]
\begin{center}
\includegraphics[width=1.0\linewidth]{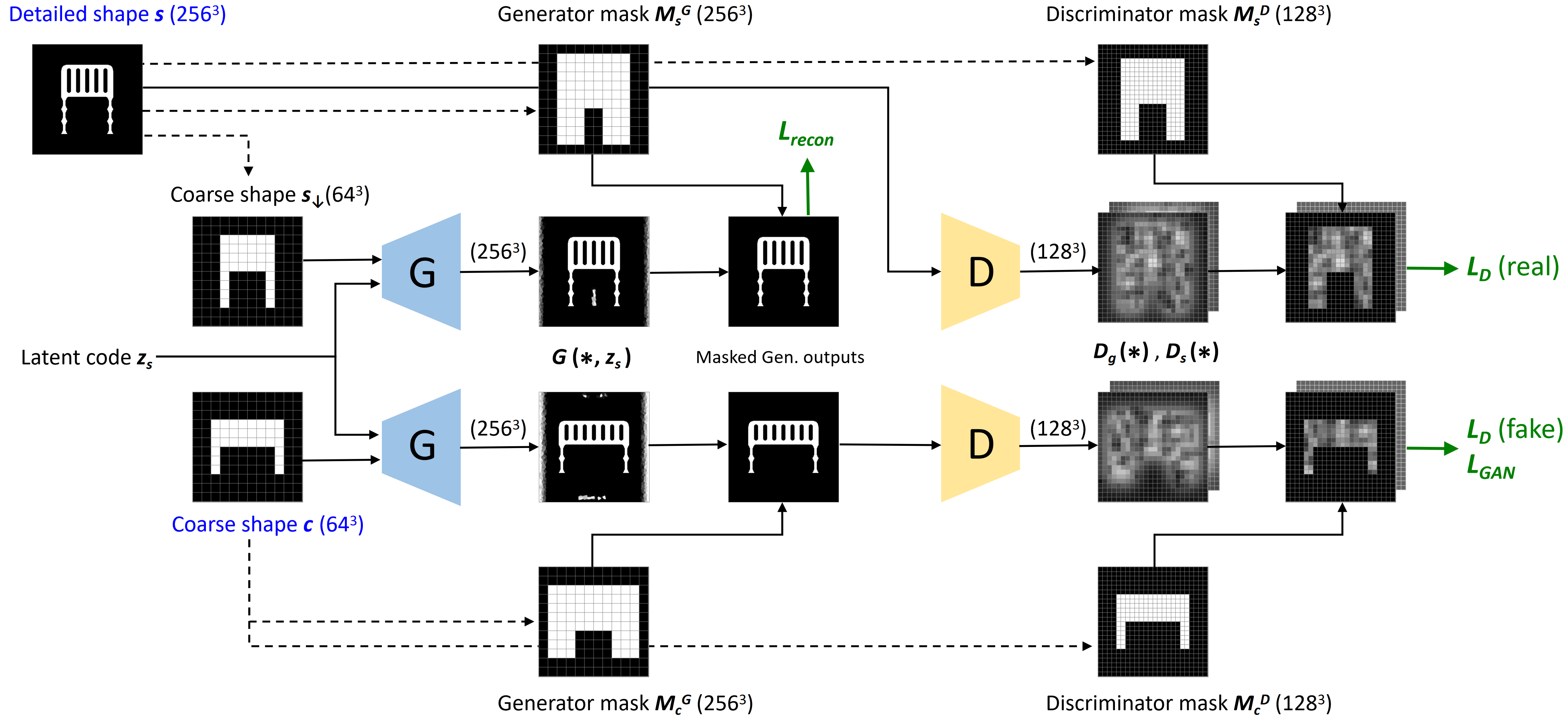}
\end{center}
\caption{
The network architecture. The training data is shown in blue and the loss functions are shown in green. Note that there is only one generator $G$ and discriminator $D$ -- the blocks are duplicated for clarity.
\vspace{-2mm}
}
\label{fig:outline_net}
\end{figure*}

%% file: 3_method.tex
\section{Method}
\label{sec:method}

\subsection{Network architecture and loss functions}
\label{sec:network}

The network structure for DECOR-GAN is shown in Figure~\ref{fig:outline_net}. In the training stage, the network receives one of the $M$ coarse shapes ($64^3$ voxels, referred to as ``content shapes''), as well as a latent code describing the style of one of the $N$ detailed shapes ($256^3$ voxels). The network is trained to upsample the content shape by 4 times according to the style of the detailed shape.

\vspace{-10pt}
\paragraph{Network overview.} We use a GAN approach and  train a generative model that can generate a shape with local patches similar to those of the detailed shapes. For the generator we utilize a 3D CNN, and for the discriminator we use 3D CNN PatchGANs~\cite{isola2017image} with receptive fields of $18^3$. We use an embedding module to learn an 8-dimensional latent style code for each given detailed shape.  Please refer to the supplementary material for the detailed architectures of the networks. Note that we only use one generator and one discriminator in our system, and Figure~\ref{fig:outline_net} shows duplicated networks solely for the sake of clarity.

\vspace{-10pt}
\paragraph{Enforcing consistency between coarse input and fine output.} We guide the generator to generate upsampled voxels that are consistent with the coarse input voxels, in the sense that the downsampled version of the fine output should be identical to the coarse input content shape. To enforce this, we employ two masks: a generator mask and a discriminator mask. 
The generator mask ensures that empty voxels in the input content shape correspond with empty voxels in the output, and enables the generator to focus its capacity solely on producing plausible voxels within the valid region. We use a dilated mask to allow the generator to have some freedom to accommodate for different styles as well as allow for more topological variations. Figure~\ref{fig:mask_effect} demonstrates the effect of using the generator mask.
The discriminator mask ensures that each occupied voxel in the input content shape leads to creation of fine voxels in its corresponding area of the output.
We apply discriminator masks on real and fake patches to only keep the signals of regions corresponding to occupied patches, so that lack of voxels in those patches will be punished. In our experiments, we use discriminator masks with 1/2 of the resolution of the detailed shapes to fit the entire model into the GPU memory. More details of generating our masks are provided in the supplementary.

\vspace{-10pt}
\paragraph{Preventing mode collapse.} Employing one global discriminator may result in mode collapse: the generator may output the same style regardless of the input style code, and the discriminator may ignore the style when determining whether a patch is plausible or not. Therefore, our discriminator is split into $N+1$ branches at the output layer, where $N$ is the number of detailed shapes used in training, with the additional $1$ standing for a global discriminator branch considering all styles. During training, the global branch computes a global GAN loss, and the appropriate style-specific branch computes a style-specific loss. The two losses are weighed in order to control the tradeoff between general plausibility and stylization.

\vspace{-10pt}
\paragraph{Loss function.} We now detail the different terms in our loss function. We denote the set of $N$ detailed shapes by $\mathcal{S}$ and the set of $M$ coarse shapes by $\mathcal{C}$. The binary discriminator masks of shape $s \in \mathcal{S}$ and shape $c \in \mathcal{C}$ are denoted as $M^D_s$ and $M^D_c$, respectively. The binary generator masks of shape $s \in \mathcal{S}$ and shape $c \in \mathcal{C}$ are denoted as $M^G_s$ and $M^G_c$, respectively. We denote the latent style code of $s \in \mathcal{S}$ by $z_s$. We denote the generator and discriminator as $G$ and $D$, respectively. The global branch of $D$ is denoted as $D_{g}$ and the branch for the detailed shape $s \in \mathcal{S}$ is denoted as $D_{s}$. $\circ$ stands for element-wise product. We use the adversarial loss in LSGAN~\cite{mao2017least} as our GAN loss, therefore, our discriminator loss is composed of the global branch's loss and the style branch's loss:
\begin{equation}
L_{D} = L_{D}^{global} + L_{D}^{style} ,
\end{equation}
where
\begin{align}
\begin{aligned}
L_{D}^{global} & = \mathbb{E}_{s \sim \mathcal{S}} \frac{ || (D_{g} (s) - 1) \circ M^D_s ||_2^2 }{||M^D_s||_1} \\
 & + \mathbb{E}_{s \sim \mathcal{S}, c \sim \mathcal{C}} \frac{ || D_{g} (G(c,z_s) \circ M^G_c) \circ M^D_c ||_2^2 }{||M^D_c||_1},
\end{aligned}
\end{align}
\begin{align}
\begin{aligned}
L_{D}^{style} & = \mathbb{E}_{s \sim \mathcal{S}} \frac{ || (D_{s} (s) - 1) \circ M^D_s ||_2^2 }{||M^D_s||_1} \\
 & + \mathbb{E}_{s \sim \mathcal{S}, c \sim \mathcal{C}} \frac{ || D_{s} (G(c,z_s) \circ M^G_c) \circ M^D_c ||_2^2 }{||M^D_c||_1} .
\end{aligned}
\end{align}
Our GAN loss for the generator is
\begin{equation}
L_{GAN} = L_{GAN}^{global} + \alpha \cdot L_{GAN}^{style} ,
\end{equation}
where
\begin{align}
\begin{aligned}
L_{GAN}^{global} = \mathbb{E}_{s \sim \mathcal{S}, c \sim \mathcal{C}} \frac{ || (D_{g} (G(c,z_s) \circ M^G_c) - 1) \circ M^D_c ||_2^2 }{||M^D_c||_1} ,
\end{aligned}
\end{align}
\begin{align}
\begin{aligned}
L_{GAN}^{style} & = \mathbb{E}_{s \sim \mathcal{S}, c \sim \mathcal{C}} \frac{ || (D_{s} (G(c,z_s) \circ M^G_c) - 1) \circ M^D_c ||_2^2 }{||M^D_c||_1} .
\end{aligned}
\end{align}
In addition, we add a reconstruction loss: if the input coarse shape and the style code stem from the same detailed shape, we expect the output of the generator to be the ground-truth detailed shape. 
\begin{equation}
L_{recon} = \mathbb{E}_{s \sim \mathcal{S}} \frac{ || G(s_{\downarrow},z_s) \circ M^G_s - s ||_2^2 }{|s|}
\end{equation}

where $|s|$ is the volumn (height $\times$ width $\times$ depth) of the voxel model $s$, and $s_{\downarrow}$ is the content shape downsampled from $s$.
The reconstruction loss both speeds up convergence in the early stage of training, as well as helps avoid mode collapse.

The final loss function for the generator is a weighted sum of the GAN loss and the reconstruction loss.
\begin{equation}
L_{total} = L_{GAN} + \beta \cdot L_{recon}
\end{equation}

\vspace{-5pt}
\subsection{Implementation details}

In our implementation, we take several measures to handle the large memory footprint incurred when processing 3D high-resolution voxels. We use $256^3$ voxels as our high-resolution detailed shapes and $64^3$ voxels as low-resolution content shapes. In order to discard unused voxels, we crop each shape according to its dilated bounding box by cropping the content shape first, then use the upsampled crop region as the crop region of the high resolution models. Moreover, since most man-made shapes are symmetric, we assume all training shapes possess bilateral symmetry, and hence only generate half of the shape. 

Another important consideration is that voxels are expected to hold binary values, as opposed to continuous intensities in images. As reported in the supplementary material of IM-NET~\cite{IMNET}, a naive GAN with CNN architectures performs poorly on binary images, as pixels with any value other than $0$ or $1$ will be considered as fake by the discriminator, thus preventing continuous optimization. We follow the approach in IM-NET and apply a Gaussian filter with $\sigma=1.0$ to the training high-resolution voxel grids to make the boundary smoother, as a pre-processing step. We provide some analysis in Sec \ref{sec:ablation}.

We set $\alpha = 0.2$ for car, $\alpha = 0.1$ for airplane, and $\alpha = 0.5$ for chair. We set $\beta = 10.0$. More discussion about the parameters can be found in Sec \ref{sec:ablation}. We train individual models for different categories for 20 epochs on one Nvidia Tesla V100 GPU; training each model takes 6 to 24 hours depending on the category.

%% file: 4_results.tex
\section{Results and evaluation}
\label{sec:results}

We conducted experiments on three categories from ShapeNet~\cite{chang2015shapenet}: car, airplane, and chair. We use only the bilaterally-symmetric models, apply an 80\%~/~20\% train/test split of the coarse content shapes, and select 64 fine detailed shapes. We choose a $64^3$ voxel resolution for coarsening airplanes and cars, and a coarser resolution of $32^3$ for chairs to further remove topological details. Note that our network is designed to upsample any input by 4 times. We use marching cubes~\cite{lorensen1987marching} to extract the surfaces visualized in the figures.
More categories (table, motorbike, laptop, and plant) can be found in the supplementary, where we lift the bilateral symmetry assumption for some categories.

\subsection{Style-content hybrids}
\input{figures/hybrid.tex}

In Figure~\ref{fig:hybrid}, we show results obtained by upsampling a content shape with a latent code to guide the style of the output shape. While the \emph{style-specific} discriminator encourages the generator to use style-appropriate patches, the \emph{global} discriminator ensures that in case no such plausible patches exist, the generator will compromise on using patches from other styles and not generate implausible details. More results can be found in the supplementary.

\subsection{Latent space}
\input{figures/latent_codes.tex}

We can explore styles in a continuous way within the $8$-dimensional latent space of styles, and have created a GUI app for it (see Sec~\ref{sec:gui}). We visualize the chair style space in Figure~\ref{fig:latent_codes}, revealing grouping of similar styles.

\subsection{Evaluation metrics}
\label{sec:evaluationmetrics}

We now discuss the metrics used to evaluate our method. See the supplementary for the full details.
\textbf{Strict-IOU} measures the Intersection over Union between the downsampled output voxels and the input voxels to evaluate how much the output respects the input. \textbf{Loose-IOU} is a relaxed version of Strict-IOU, which computes the percentage of occupied voxels in the input that are also occupied in the downsampled output. Local Plausibility (LP) is defined as the percentage of local patches in the output shape that are ``similar'' (according to IOU or F-score) to at least one local patch in the detailed shapes, which defines the metrics \textbf{LP-IOU} and \textbf{LP-F-score}. We evaluate the Diversity (Div) of the output shapes by computing the percentage of output shapes that are consistent with their input style code. An output shape with input style code $z_s$ whose local patches are most similar to those of the detailed shape $s$ is considered as a ``consistent'' output. We measure \textbf{Div-IOU} and \textbf{Div-F-score} according to similarity metrics for patches. We evaluate the plausibility of the generated shapes by training a classifier to distinguish generated and real shapes, and use the mean classification accuracy as \textbf{Cls-score}. Following Fr\'echet Inception Distance (FID) ~\cite{heusel2017gans}, we use FID to compare the output shapes with all available shapes in a category, denoted as \textbf{FID-all}; or only the detailed shapes providing styles, denoted as \textbf{FID-style}.
For Cls-score and FID, a lower score is better; for others, a higher score is better.

\subsection{Ablation study}
\label{sec:ablation}

We now verify the necessity of the various parts of our network. We report the quantitative results for chairs in this section, and other categories in the supplementary.

\vspace{-10pt}
\paragraph{Generator and Discriminator Masks.}

\input{figures/ablation.tex}

\begin{table*}[!t]
\begin{center}
\resizebox{1.0\linewidth}{!}{
\begin{tabular}{l|c|c|c|c|c|c|c|c|c}
\hline
  & Strict-IOU $\uparrow$ & Loose-IOU $\uparrow$ & LP-IOU $\uparrow$ & LP-F-score $\uparrow$ & Div-IOU $\uparrow$ & Div-F-score $\uparrow$ & Cls-score $\downarrow$ & FID-all $\downarrow$ & FID-style $\downarrow$ \\
\hline
Recon. only         & \bf{0.976} & \bf{0.993} & 0.260 & 0.935 & 0.325 & 0.188 & 0.627 & 53.2 & 411.7 \\
No Gen. mask        & 0.655 & 0.792 & \bf{0.452} & \bf{0.973} & \bf{0.825} & 0.806 & 0.672 & 121.9 & 379.9 \\
Strict Gen. mask    & 0.587 & 0.587 & 0.344 & 0.941 & 0.150 & 0.100 & 0.750 & 305.5 & 548.2 \\
No Dis. mask        & 0.145 & 0.167 &  N/A  &  N/A  &  N/A  &  N/A  & 0.843 & 2408.9 & 2714.1 \\
Conditional Dis. 1  & 0.947 & 0.981 & 0.259 & 0.949 & 0.291 & 0.194 & \bf{0.593} & \bf{51.3} & 402.7 \\
Conditional Dis. 3  & 0.928 & 0.977 & 0.246 & 0.963 & 0.197 & 0.206 & 0.603 & 55.8 & 418.2 \\
Proposed method     & 0.673 & 0.805 & 0.432 & \bf{0.973} & 0.800 & \bf{0.816} & 0.644 & 113.1 & \bf{372.5} \\
\hline
\end{tabular}
}
\end{center}
\caption{Ablation on the generator and discriminator masks. ``N/A'' is due to empty outputs. Best results are marked in bold.}
\label{table:Ablation_mask}
\end{table*}

In Table~\ref{table:Ablation_mask}, and Figure~\ref{fig:ablation} (top) we compare our proposed method with several variations on it: {\bf a.} {\em Recon. only}, in which we remove the discriminator and train the network with only $L_{recon}$, to validate the effectiveness of adversarial training. This results in the network essentially mode-collapsing (reflected by the low Div scores). {\bf b.} {\em No Gen. mask}, in which we remove the generator mask. To still respect the input shape, we add a loss term to punish any voxels generated outside the generator mask. This results in comparable performance to our method locally, but since the generator needs to allocate a portion of its capacity to ensure no voxels are generated outside the generator mask, it is left with less capacity for generating the output shape.  This is reflected in deterioration of global plausibility (reflected by the higher Cls-score). {\bf c.} {\em Strict Gen. mask}, in which we use the un-dilated, ``strict'' generator mask, which results in a harsh drop in quality. {\bf d.} {\em No Dis. mask}, in which we remove the discriminator mask, performs even worse, with some outputs completely empty, because patches with no occupied voxels are considered as real by the discriminator. {\bf e.} {\em Conditional Dis. 1}, in which we remove both the generator mask and the discriminator mask, and condition the discriminator on the occupancy of the input coarse voxels, i.e., a conditional GAN. In this setting, each discriminator output will have a receptive field of $1^3$ in the input coarse voxels. This leads to failure in generating diverse outputs, possibly due to the discriminator wasting capacity on parsing the input conditions. {\bf f.} {\em Conditional Dis. 3} modifies the receptive field of {\em Conditional Dis. 1} to $3^3$, but runs into similar issues.

One may notice that there is a considerable difference between LP-IOU and LP-F-score, and they sometimes contradict  each other. This is due to IOU being sensitive to small perturbations on the surface, especially when the structure is thin. F-score, on the other hand, is less sensitive. Due to the strictness of IOU, Div-IOU is usually higher than Div-F-score. In addition, we observe that Cls-score is consistent with apparent visual quality. Therefore, in the following, we  only report Loose-IOU, LP-F-score, Div-IOU and Cls-score. The full tables can be found in supplementary.

We also show the effect of the generator mask in Figure~\ref{fig:mask_effect}. Evidently, the raw generator output (b) has various artifacts outside the masked region, which are removed upon applying the generator mask (a). However, new artifacts may be created in the process, such as the one shown in (c). Finding the connected components of (c) from (b) can remove such artifacts.

\vspace{-10pt}
\paragraph{Parameter $\alpha$ and the number of styles.}

\begin{table}[!t]
\begin{center}
\resizebox{1.0\linewidth}{!}{
\begin{tabular}{l|c|c|c|c}
\hline
\multicolumn{5}{c}{16 detailed shapes as styles} \\
\hline
  & Loose-IOU $\uparrow$ & LP-F-score $\uparrow$ & Div-IOU $\uparrow$ & Cls-score $\downarrow$ \\
\hline
$\alpha = 0.0$                 & \bf{0.840} & 0.956 & 0.147 & 0.695 \\
$\alpha = 0.2$                 & 0.750 & \bf{0.971} & 0.875 & \bf{0.667} \\
$\alpha = 0.5$                 & 0.738 & 0.970 & 0.997 & 0.690 \\
No \small{$L_{GAN}^{global}$}  & 0.735 & 0.963 & \bf{1.000} & 0.692 \\
\hline
\multicolumn{5}{c}{32 detailed shapes as styles} \\
\hline
$\alpha = 0.0$                 & \bf{0.864} & 0.962 & 0.184 & \bf{0.598} \\
$\alpha = 0.2$                 & 0.812 & \bf{0.974} & 0.838 & 0.636 \\
$\alpha = 0.5$                 & 0.757 & \bf{0.974} & 0.934 & 0.662 \\
No \small{$L_{GAN}^{global}$}  & 0.728 & 0.969 & \bf{0.997} & 0.690 \\
\hline
\multicolumn{5}{c}{64 detailed shapes as styles}\\
\hline
$\alpha = 0.0$                 & \bf{0.868} & 0.983 & 0.163 & \bf{0.589} \\
$\alpha = 0.2$                 & 0.864 & \bf{0.985} & 0.353 & 0.619 \\
$\alpha = 0.5$                 & 0.805 & 0.973 & 0.800 & 0.644 \\
No \small{$L_{GAN}^{global}$}  & 0.741 & 0.965 & \bf{0.950} & 0.669 \\
\hline
\end{tabular}
}
\end{center}
\caption{Ablation on parameter $\alpha$ and the number of styles.}
\label{table:Ablation_alpha}
\end{table}

In Tables~\ref{table:Ablation_alpha} we show results of setting the parameter $\alpha$ to values $0.0, 0.2, 0.5$, as well as completely omitting $L_{GAN}^{global}$, which can be seen as using a very large $\alpha$ that would dominate the other terms. $\alpha$ controls the trade-off between global plausibility (and respecting the coarse content), and adherence to the style code. As shown by the increase in Div-IOU, the higher $\alpha$ is, the more stylized the output will be. However, the decrease in Loose-IOU and increase in Cls-score hint that a higher $\alpha$ also makes the output less considerate of the content and less globally plausible. This can also be seen in the qualitative results in Figure~\ref{fig:ablation} (bottom).

Simultaneously, we also test the effect of varying the size of the detailed-shape dataset used in training, between 16, 32, and 64. As expected, both Cls-score and Loose-IOU monotonically improve with the increase in dataset size, showing plausibility improves, because the network has a larger collection of patches to choose from. This can be seen in Figure~\ref{fig:ablation} (g), where the network was trained with 16 detailed shapes, and generated small bumps on the seat and the back, clearly stemming from a vestige of the arm. The same phenomenon can be found in (l) where there is no global discriminator. By increasing the size of the detailed-shape dataset, the vestige disappears and the back becomes more square-like, as shown in (g) (h) (k).

\vspace{-10pt}
\paragraph{Gaussian filter.}

\input{figures/mask_effect.tex}
\input{figures/sigma.tex}

\begin{table}[!t]
\begin{center}
\resizebox{1.0\linewidth}{!}{
\begin{tabular}{l|c|c|c|c}
\hline
  & Loose-IOU $\uparrow$ & LP-F-score $\uparrow$ & Div-IOU $\uparrow$ & Cls-score $\downarrow$ \\
\hline
$\sigma = 0.0$  & \bf{0.952} & 0.943 & 0.153 & \bf{0.544} \\
$\sigma = 0.5$  & 0.919 & 0.952 & 0.172 & 0.580 \\
$\sigma = 1.0$  & 0.805 & 0.973 & 0.800 & 0.644 \\
$\sigma = 1.5$  & 0.719 & \bf{0.985} & \bf{0.944} & 0.667 \\
$\sigma = 2.0$  & 0.614 & 0.982 & 0.575 & 0.711 \\
\hline
\end{tabular}
}
\end{center}
\caption{
Ablation on $\sigma$ of the Gaussian filter.
\vspace{-2mm}
}
\label{table:Ablation_sigma}
\end{table}

The effect of the preprocessing with the Gaussian filter blurring of the detailed shapes is shown in Table~\ref{table:Ablation_sigma} and  Figure~\ref{fig:sigma}. The larger $\sigma$ is, the more blurry the training high-resolution voxels are, and the better the optimization goes. Without the Gaussian filter ($\sigma=0.0$), the output looks like (a) {\em Recon. only} in Figure~\ref{fig:ablation}, indicating that the network may have reached a state where it is not easy to optimize a local patch to other styles, because the patches of different styles are not continuous for the optimization. As  $\sigma$ is increased, the output shape becomes more stylized, however above some threshold, the shape becomes too blurry to be reconstructed, especially  thin structures. Progressive training, where the network is trained with larger $\sigma$ then switches to smaller and smaller $\sigma$s during multiple steps, may work better.

\subsection{GUI application}
\label{sec:gui}

As an application of our method, we have created a GUI application where a user can explore the style space, to facilitate interactive design. Please refer to the supplementary.

\subsection{GAN application}

Since state-of-the-art 3D GANs are unable to generate detailed outputs, our method can be used directly to upsample a GAN-generated shape into a detailed shape, as long as the GAN output can be converted into a voxel grid. In Figures~\ref{fig:IMGAN}, we show results on upsampling shapes from IM-GAN~\cite{IMNET}. See more results in the supplementary.

%% file: figures/hybrid.tex
\begin{figure*}[t!]
\begin{center}
\includegraphics[width=1.0\linewidth]{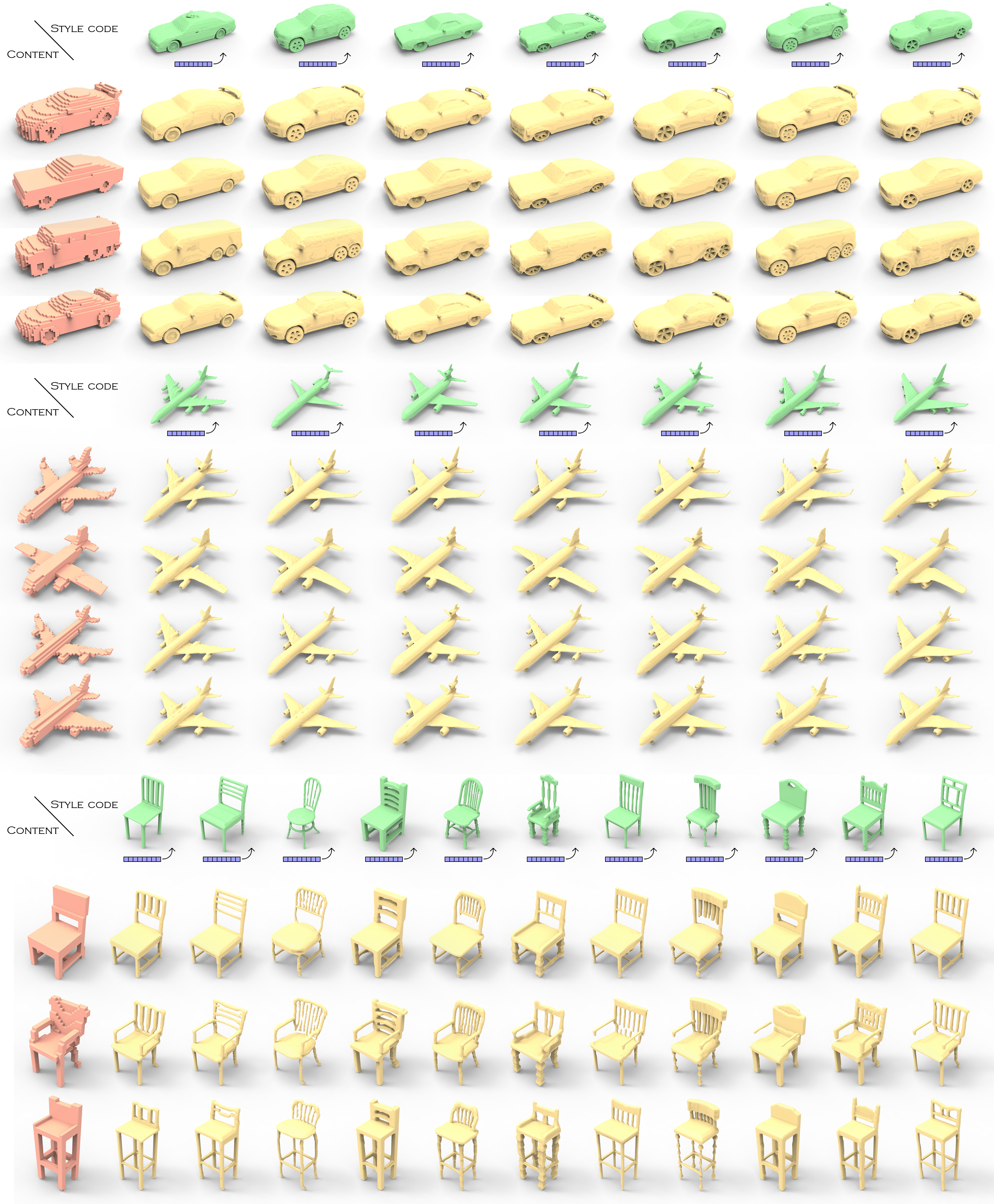}
\end{center}
\caption{
Results by upsampling coarse voxels with different style codes. In each table, we show on the top the detailed shapes that correspond to the input style codes. We show the input coarse voxel models on the left, where chairs are $32^3$ and the others are $64^3$.
}
\label{fig:hybrid}
\end{figure*}

%% file: figures/latent_codes.tex
\begin{figure}[t!]
\begin{center}
\includegraphics[width=1.0\linewidth]{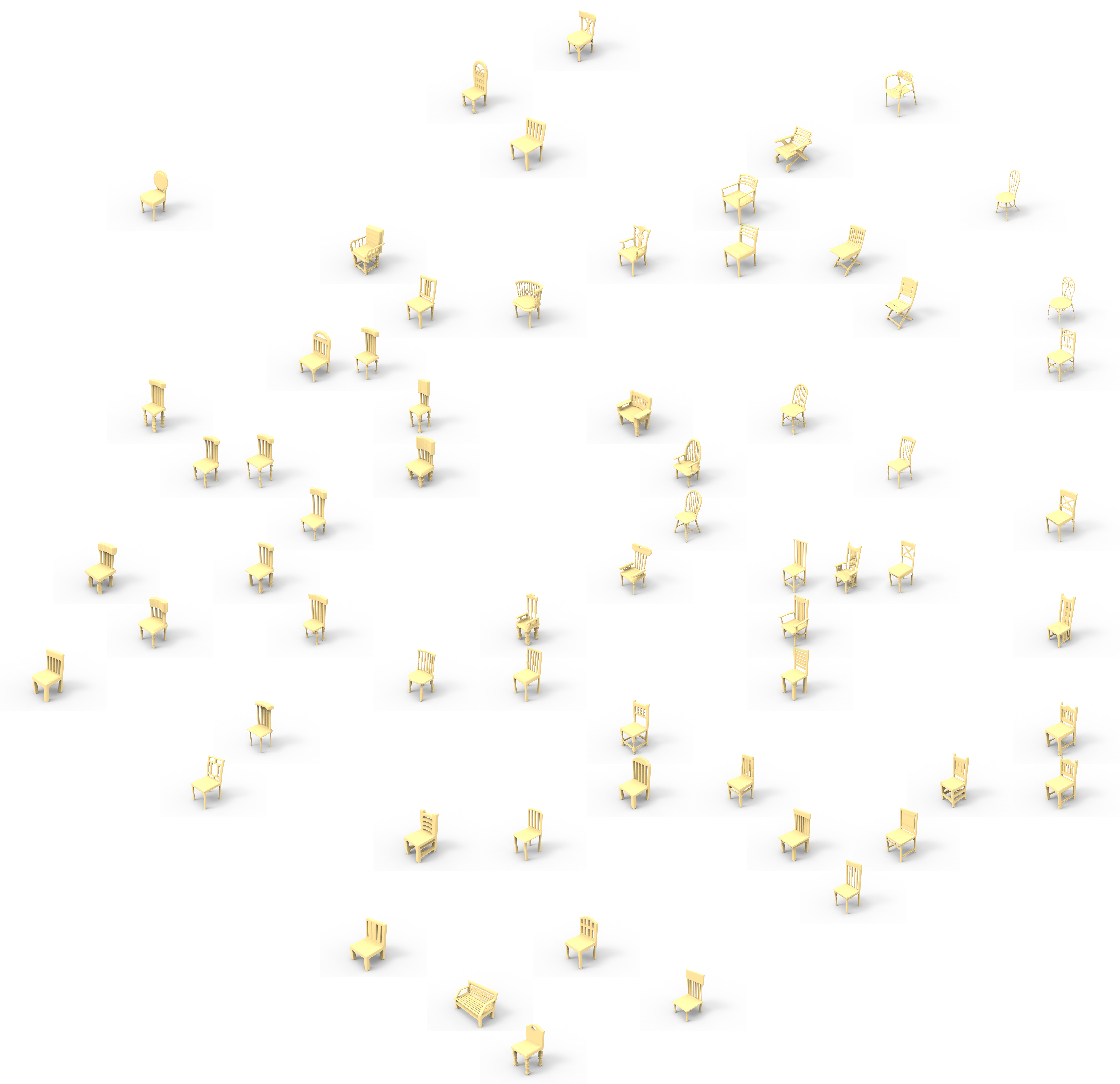}
\end{center}
\caption{
Visualization of 64 latent codes for chairs via T-SNE embedding. For each latent code, the corresponding style shape is displayed in its location.
}
\label{fig:latent_codes}
\end{figure}

%% file: figures/ablation.tex
\begin{figure}[t!]
\begin{center}
\includegraphics[width=1.0\linewidth]{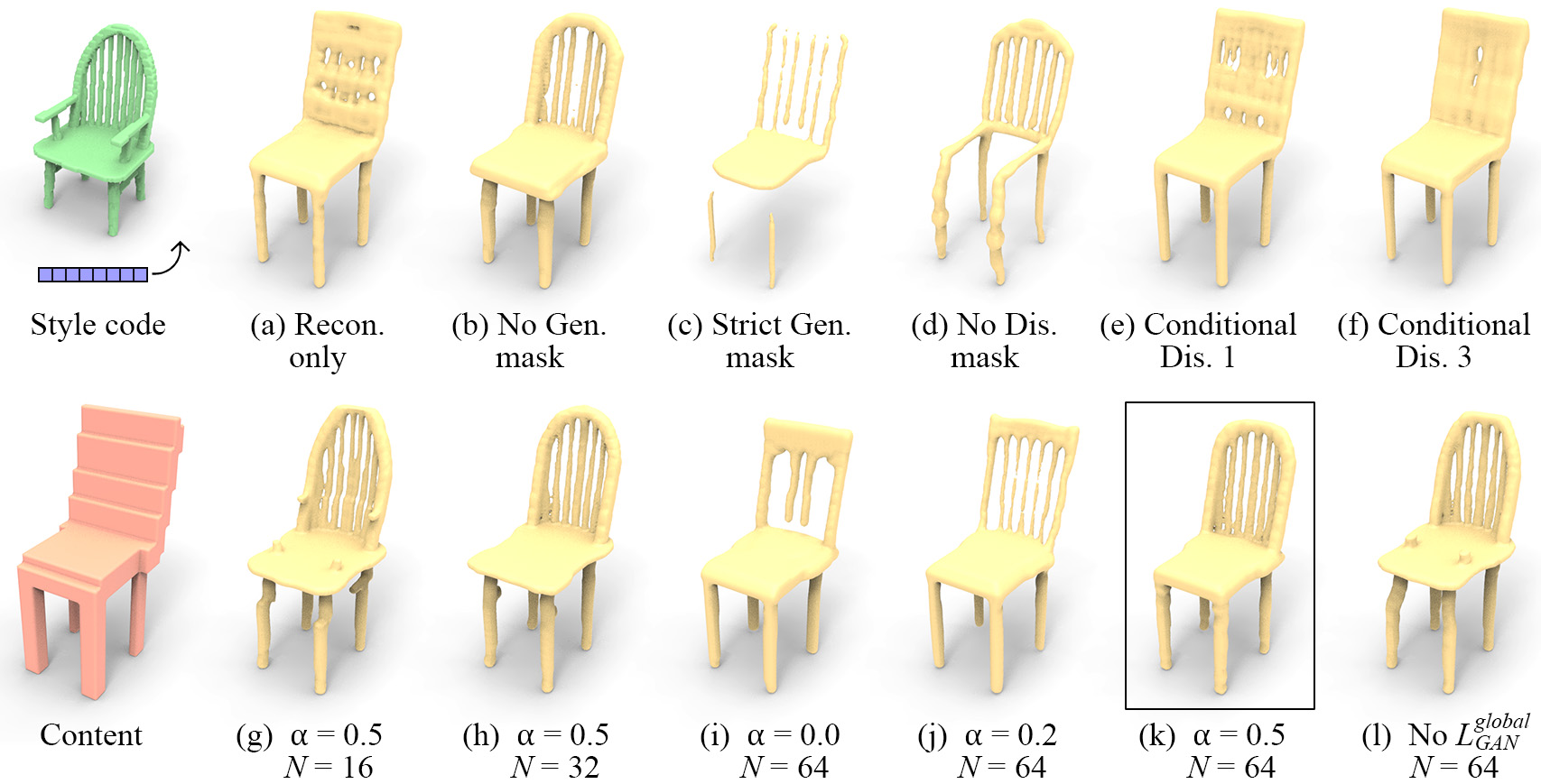}
\end{center}
\caption{
Ablation study. The input content shape and style code are shown on the left. The result with our default parameters is placed inside a box.
\vspace{-2mm}
}
\label{fig:ablation}
\end{figure}

%% file: figures/mask_effect.tex
\begin{figure}[t!]
\begin{center}
\includegraphics[width=1.0\linewidth]{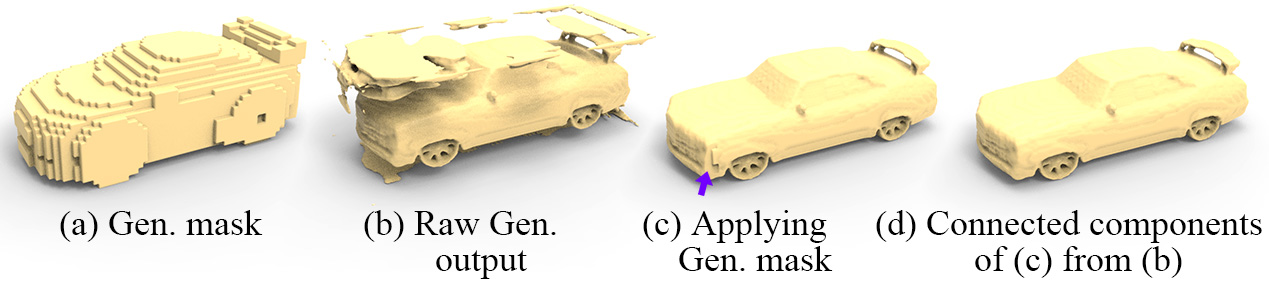}
\end{center}
\caption{
The effectiveness of the generator mask. This example is taken from row 2, column 5 of Figure~\ref{fig:hybrid}.
\vspace{-2mm}
}
\label{fig:mask_effect}
\end{figure}

%% file: figures/sigma.tex
\begin{figure}[t!]
\begin{center}
\includegraphics[width=1.0\linewidth]{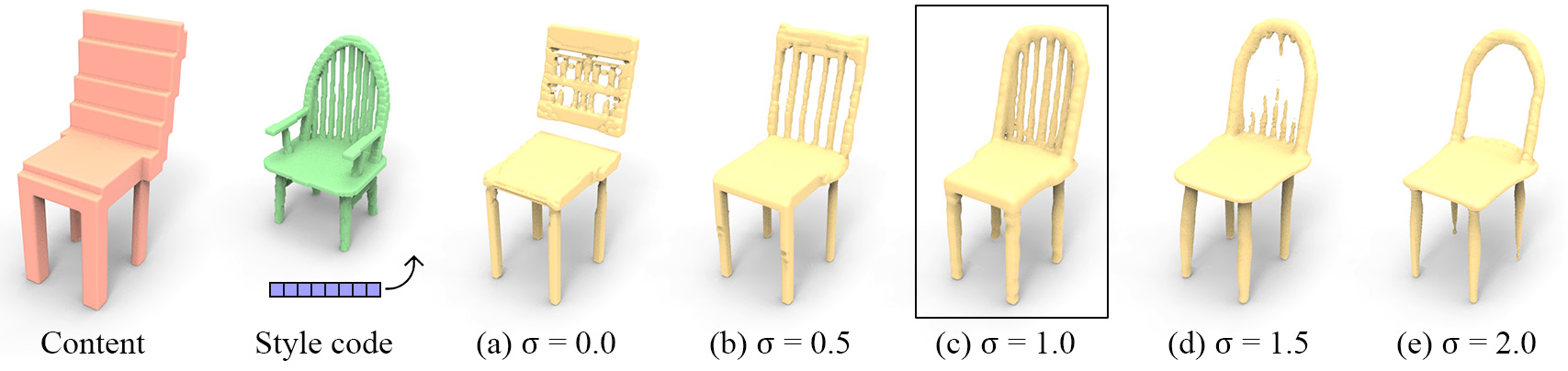}
\end{center}
\caption{
Ablation study on $\sigma$ of the Gaussian filter. The input content shape and style code are shown on the left. The result with default parameters is placed inside a box.\vspace{2mm}
}
\label{fig:sigma}
\end{figure}

%% file: 5_future.tex
\section{Conclusion, limitation, and future work}
\label{sec:future}

\input{figures/IMGAN.tex}
\input{figures/limitationfig.tex}

This paper introduces the first method to perform high-resolution conditional detail-generation for coarse 3D models represented as voxels. The coarse input enables control over the general shape of the output, while the input style code enables control over the type of details generated, which in tandem yield a tractable method for generating plausible variations of various objects, either driven by humans or via automatic randomized algorithms.

One main limitation is memory, similar to other voxel methods: a high-resolution voxel model, e.g., $256^3$, overflows GPU memory when upsampled to $1024^3$. We would like to explore more local networks to upsample patch-by-patch, or a recursive network. A second limitation is that we mainly transfer local patches from the training shapes to the target. Therefore, we cannot generate unseen patterns, e.g., a group of slats on a chair back with a different frequency from those on training shapes, see last row of Figure~\ref{fig:IMGAN}.
The network also lacks awareness to global structures and in some cases the generated shapes may present topological inconsistencies, see, e.g., second-to-last row of Figure~\ref{fig:hybrid}.
Lastly, as is often the case with GAN training, parameter tuning may be elaborate and fragile, see Figure~\ref{fig:limitationfig}.

Many immediate follow-ups suggest themselves. One would be to learn a meaningful, smooth, latent space, so that all latent codes will produce valid styles and latent space interpolation produce smooth transitions. Likewise, exploring hierarchies of details could enable more elaborate and consistent output. A complete decoupling of fine details, the coarse shape, and the semantic shape category is also interesting as it would enable training across larger collections of shapes, with the same styles employed across different shape categories. Lastly, we of course eye various advancements in voxel representation to reach higher resolutions.

We are excited about the future prospects of this work for detailization of 3D content. Immediate applications include amplification of stock data, image-guided 3D style generation, and enabling CAD-like edits that preserve fine detail.

%% file: figures/IMGAN.tex
\begin{figure}[t!]
\begin{center}
\includegraphics[width=1.0\linewidth]{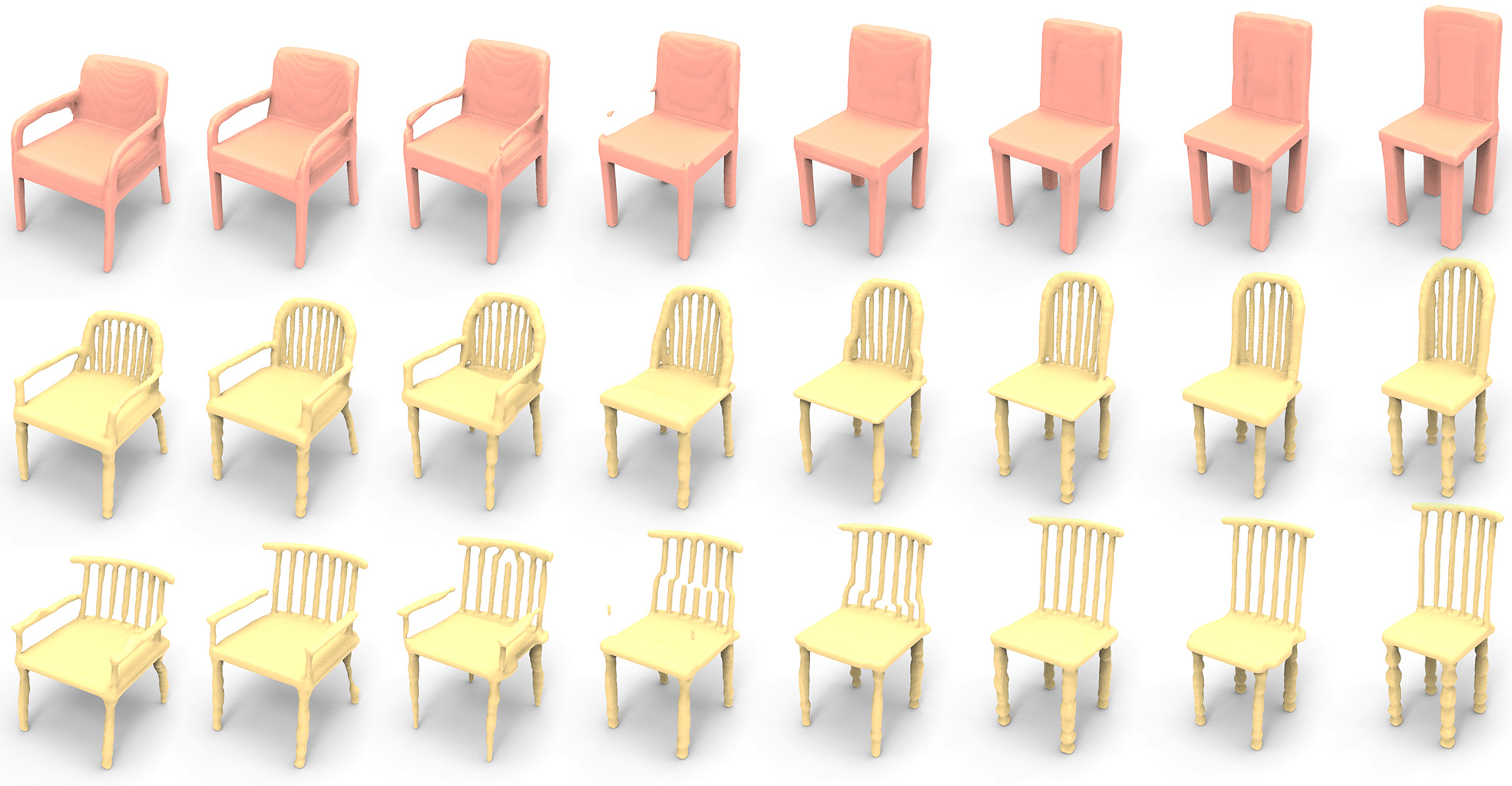}
\end{center}
\caption{
Upsampling GAN-generated shapes into detailed shapes. In the first row we show a sequence of generated shapes from IM-GAN via linearly interpolating two random latent codes. The last two rows show our upsampled results.
}
\label{fig:IMGAN}
\end{figure}

%% file: figures/limitationfig.tex
\begin{figure}[t!]
\begin{center}
\includegraphics[width=1.0\linewidth]{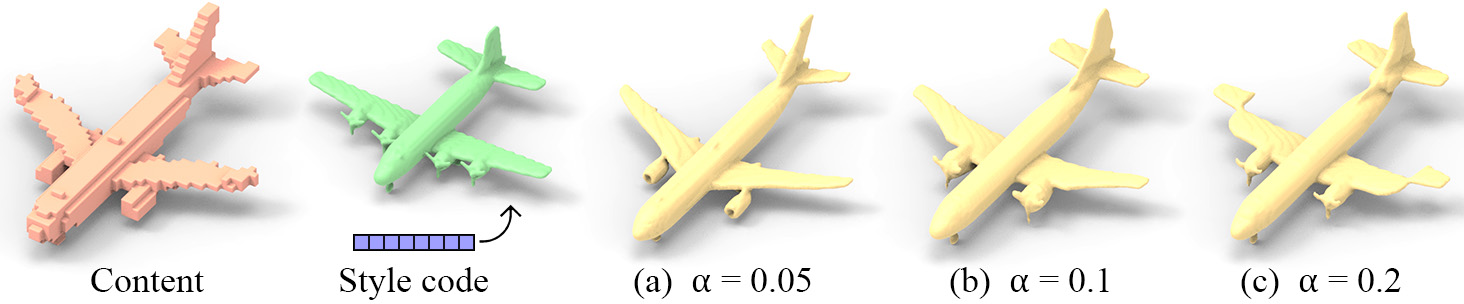}
\end{center}
\caption{
Sensitivity on parameter $\alpha$. The input content shape and style code are shown on the left.
\vspace{-2mm}
}
\label{fig:limitationfig}
\end{figure}

%% file: 6_acks.tex
\vspace{-10pt}

\paragraph{Acknowledgements.}
We thank the anonymous reviewers for their valuable comments.
This work was completed while the first author was carrying out an internship at Adobe Research; it is
supported in part by an NSERC grant (No.~611370) and a gift fund from Adobe.

%% file: 6_supplementary.tex
\appendix
\section{Supplementary material}

\subsection{Details of network architectures}
We provide detailed architectures of the generator and the discriminator used in our system in Figure ~\ref{fig:network_detail}. We train individual models for different categories for 20 epochs on one Nvidia Tesla V100 GPU. Training each model takes approximately 6 hours for category chair, 12 hours for airplane, and 24 hours for car. The training batch size is set to 1. We use Adam optimizer with lr=0.0001, beta1=0.9, beta2=0.999.

\subsection{The generator and discriminator masks}

First, To ensure each empty voxel in the input content shape leads to empty voxels in its corresponding area of the output, we mask out voxels generated outside a predefined valid region. The region is denoted as the generator mask. There are two masking options: the ``strict'' generator mask, by upsampling the occupied voxels in the content shape to the desired resolution; and the ``loose'' generator mask, by upsampling the occupied voxels in the content shape after dilating them by 1 voxel. In both cases we use nearest-neighbor upsampling. In our system, we apply the ``loose'' generator mask to the raw generator output to keep only voxels within the area of the mask. The reason for using the ``loose'' mask is to allow the generator to have some freedom to accommodate for different styles as well as allow for more topological variations as the dilation may close holes. The generator mask enables the generator to focus its capacity solely on producing plausible voxels within the valid region.

Second, to ensure each occupied voxel in the input content shape leads to creation of fine voxels in its corresponding area of the output, we require that an occupied coarse voxel is also occupied in the downsampled version of the generator output. We achieve this by training the discriminator to penalize lack of voxels. If all real patches used in training have at least one voxel occupied at their center $4^3$ areas, then any patches that have empty $4^3$ center areas will be considered fake under the view of the discriminator. Therefore, the discriminator will encourage all input patches to have occupied voxels in their center areas. Hence, we can encourage voxels to be generated inside the desired region by {\em a.} training the discriminator using patches with occupied center areas as real patches, and {\em b.} training the generator by feeding to the discriminator those local patches that should have their center areas occupied. These two can be done easily by applying binary masks to the discriminator to only keep the signals of the desired patches. For the real patches, given a detailed shape, we can obtain a {\em discriminator mask} by checking each local patch for whether their center areas are occupied by at least one voxel. For the fake (generated) shape, we obtain its {\em discriminator mask} by upsampling the content shape via nearest-neighbor. In our experiments, we use discriminator masks with 1/2 of the resolution of the detailed shapes so that the entire model can fit into the GPU memory.

\subsection{Style-content hybrids}
We show more results of style-content hybrid shapes in Figure ~\ref{fig:supp_chair} ~\ref{fig:supp_car} ~\ref{fig:supp_plane} ~\ref{fig:supp_table} ~\ref{fig:supp_motor} ~\ref{fig:supp_laptop} ~\ref{fig:supp_plant}. Note that we lift the bilateral symmetry assumption for category motorbike, laptop, and plant.

\subsection{Latent space}
We show a visualization of the style space for airplanes in Figure~\ref{fig:latent_plane} and cars in Figure~\ref{fig:latent_car}. The visualization for chairs can be found in the main paper.

\subsection{Evaluation metrics}

To quantitatively evaluate the quality of the generated shapes, we propose the following metrics.

\paragraph{Strict-IOU and Loose-IOU. (higher better)} Ideally, the downsampled version of a generator output should be identical to the input content shape. Therefore, we can use the IOU (Intersection over Union) between the downsampled voxels and the input voxels to evaluate how much the output shape respects the input. We use max-pooling as the downsampling method, and the Strict-IOU is defined as described above. However, since we relaxed the constraints (see Sec 3.1 of the main paper) so that the generator is allowed to generate shapes in a dilated region, we define Loose-IOU as a relaxed version of IOU to ignore the voxels in the dilated portion of the input:
\begin{equation}
\textrm{Loose-IOU} = \frac{ | V_{in} \cap (V_{out} \cap V_{in}) | }{ | V_{in} \cup (V_{out} \cap V_{in}) | }  = \frac{ | V_{in} \cap V_{out} | }{ | V_{in} | }.
\end{equation}
where $V_{in}$ and $V_{out}$ are input voxels and downsampled output voxels, and $| V |$ counts the number of occupied voxels in $V$. Note that our generated shape is guaranteed to be within the region of the dilated input due to the generator mask.

\paragraph{LP-IOU and LP-F-score (higher better).} If all local patches from an output shape are copied from the given detailed shapes, it is likely that the output shape looks plausible, at least locally. Therefore, we define the Local Plausibility (LP) to be the percentage of local patches in the output shape that are ``similar'' to at least one local patch in the detailed shapes. Specifically, we define the distance between two patches to be their IOU or F-score. For LP-IOU, we mark the two patches as ``similar'' if the IOU is above $0.95$; for LP-F-score, we mark ``similar'' if the F-score is above $0.95$. The F-score is computed with a distance threshold of 1 (voxel). In our experiments, we sample $12^3$ patches in a voxel model. The patch size is a bit less than the receptive field of our discriminator to reduce computational complexity. In addition, we want to avoid sampling featureless patches that are mostly inside or outside the shape, therefore we only sample surface patches that have at least one occupied voxel and one unoccupied voxel at their center $2^3$ areas. We sample 1000 patches in each testing shape, and compare them with all possible patches in the detailed shapes.

\paragraph{Div-IOU and Div-F-score (higher better).} For the same input shape, different style codes should produce different outputs respecting the styles. Therefore, we want to have a metric that evaluates the diversity of the outputs with respect to the styles. During the computation of the LP, we obtain $N_{ijk}$, the number of local patches from input $i$, upsampled with style $j$, that are ``similar'' to at least one patch in detailed shape $k$. In an ideal case, any input $i$ upsampled with style $j$ only copies patches from detailed shape $j$, therefore we have $j = \max_k N_{ijk}$. However, since the input shape might introduce style bias (e.g., a local structure that can only be found in a specific detailed shape), we denote $N_{ik}$ to be the mean of $N_{ijk}$ over all possible $j$, and use it to remove such bias. The diversity is defined as
\begin{equation}
\textrm{Div} = \mathbb{E}_{i,j} [ \mathbbm{1} (j = \textrm{argmax}_k (N_{ijk} - N_{ik}) ) ].
\end{equation}
We obtain Div-IOU and Div-F-score based on the distance metrics for patches.

\paragraph{Cls-score (lower better).} If the generated shapes are indistinguishable from real samples, a well-trained classification network will not be able to classify whether a shape is real or fake. We can evaluate the plausibility of the generated shapes by training such a network and inspect the classification score. However, the network may easily overfit if we directly input 3D voxel models, since we have limited amount of real data. Therefore, we opt to use rendered images for this task. We train a ResNet~\cite{he2016deep} using high-resolution voxels (from which content shapes are downsampled) as real samples, and our generated shapes as fake samples. The samples are rendered to obtain 24 $256^2$ images from random views. The images are randomly cropped to 10 $64^2$ small patches and feed into the network for training. We use the mean classification accuracy as the metric for evaluating plausibility, denoted as Cls-score.

\paragraph{FID-all and FID-style (lower better).} Since our method generates shapes for a single category, it is not well suited for evaluation with Inception Score~\cite{salimans2016improved}. However, we borrow the idea from Fr\'echet Inception Distance (FID)~\cite{heusel2017gans} and propose a customized FID as follows. We first train a 3D CNN classification network on ShapeNet with $128^3$ or $256^3$ voxels depending on the input resolution. Afterwards, we use the last hidden layer (512-d) as the activation features for computing FID. We use FID to compare our generated shapes with all high-resolution voxels from which content shapes are downsampled, denoted as FID-all; or with a group of detailed shapes, denoted as FID-style.

\paragraph{Evaluation details} For LP and Div, we evaluate on 320 generated shapes (20 contents $\times$ 16 styles) since they are computationally expensive. For other metrics we evaluate on 1600 generated shapes (100 contents $\times$ 16 styles). We evaluate Div and FID-style with the first 16 styles, and LP with all 64 styles.

\subsection{Ablation study}
We provide all quantitative results for our ablation experiments in this section. The numbers for chairs can be found in Table~\ref{table:Ablation_chair}. The numbers for cars can be found in Table~\ref{table:Ablation_car}. The numbers for airplanes can be found in Table~\ref{table:Ablation_plane}.

\subsection{GUI application}

The video is available at \href{https://youtu.be/xIQ0aslpn8g}{https://youtu.be/xIQ0aslpn8g}.
We obtain the 2D style space via T-SNE embedding. Afterwards, we consider each style as a 2D point and obtain the Delaunay triangulation of the 2D style space. The 8D latent style code for a given 2D point can be computed by finding which triangle it is inside and compute a linear interpolation among the three 8D latent codes of the three vertices via barycentric coordinates.

\input{figures/network_detail.tex}
\input{figures/supp_chair.tex}
\input{figures/supp_car.tex}
\input{figures/supp_plane.tex}
\input{figures/supp_table.tex}
\input{figures/supp_motor.tex}
\input{figures/supp_laptop.tex}
\input{figures/supp_plant.tex}
\input{figures/latent_plane.tex}
\input{figures/latent_car.tex}
\input{tables/numbers_chair.tex}
\clearpage
\newpage
\input{tables/numbers_car.tex}
\clearpage
\newpage
\input{tables/numbers_plane.tex}

%% file: figures/network_detail.tex
\begin{figure*}[t!]
\begin{center}
\includegraphics[width=1.0\linewidth]{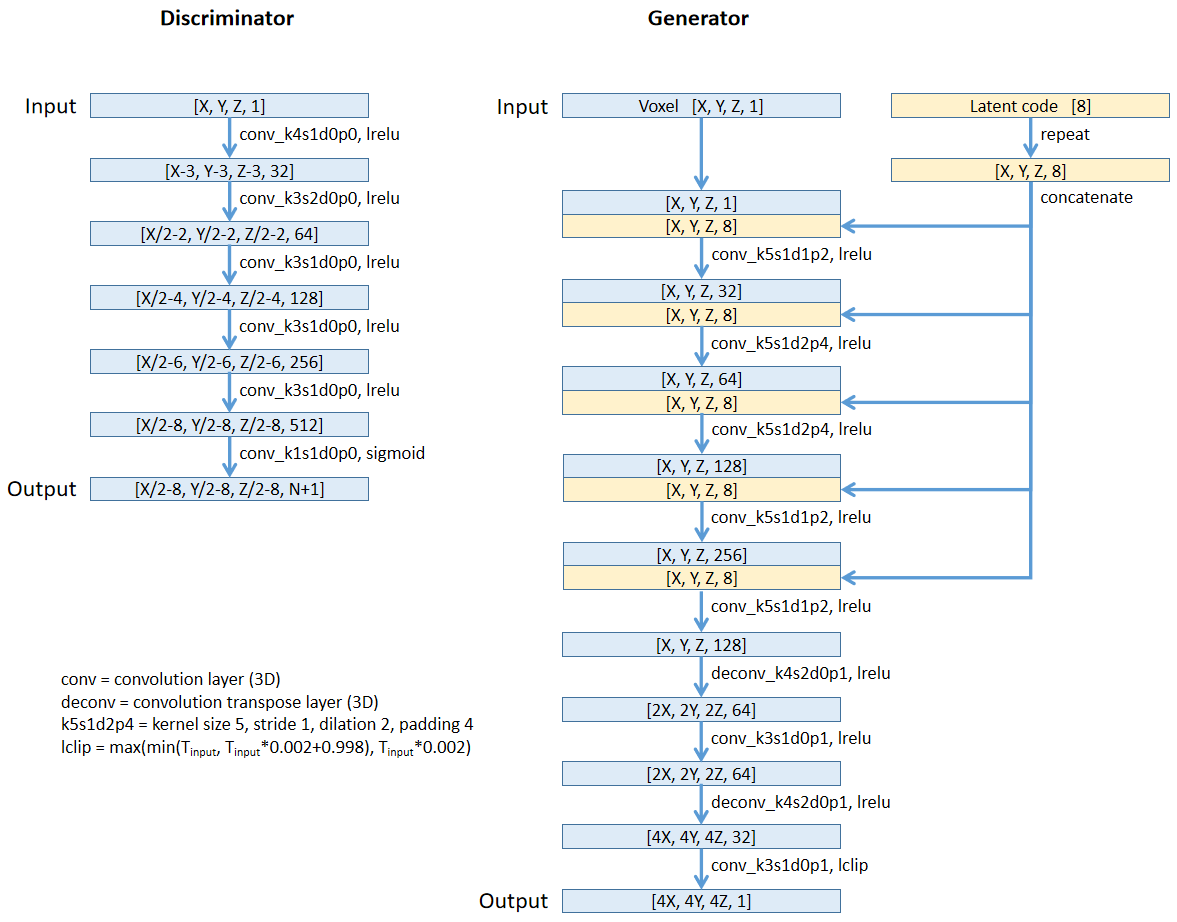}
\end{center}
\caption{
The detailed network architectures. Note that the generator for category chair with $32^3$  inputs has smaller receptive fields by replacing all kernel-5 convolution layers with kernel-3 convolution layers.
}
\label{fig:network_detail}
\end{figure*}

%% file: figures/supp_chair.tex
\begin{figure*}[t!]
\begin{center}
\includegraphics[width=1.0\linewidth]{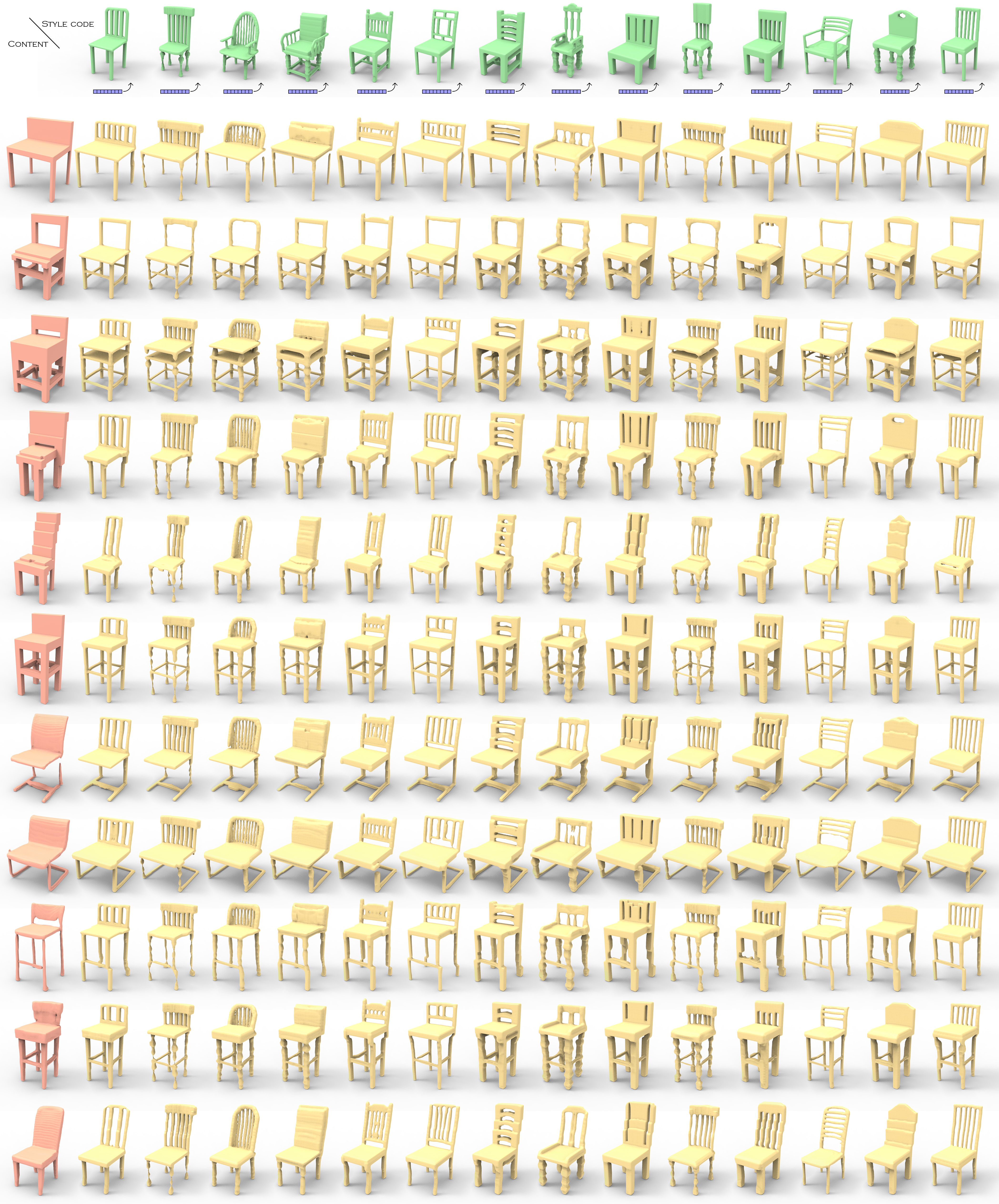}
\end{center}
\caption{
Results by upsampling coarse chairs with different style codes. We show on the top the detailed shapes that correspond to the input style codes. The input shapes are coarse voxels in the first 6 rows, and downsampled versions of shapes generated by IM-GAN~\cite{IMNET} in the last 5 rows. The input resolution is $32^3$ and the output resolution is $128^3$.
}
\label{fig:supp_chair}
\end{figure*}

%% file: figures/supp_car.tex
\begin{figure*}[t!]
\begin{center}
\includegraphics[width=1.0\linewidth]{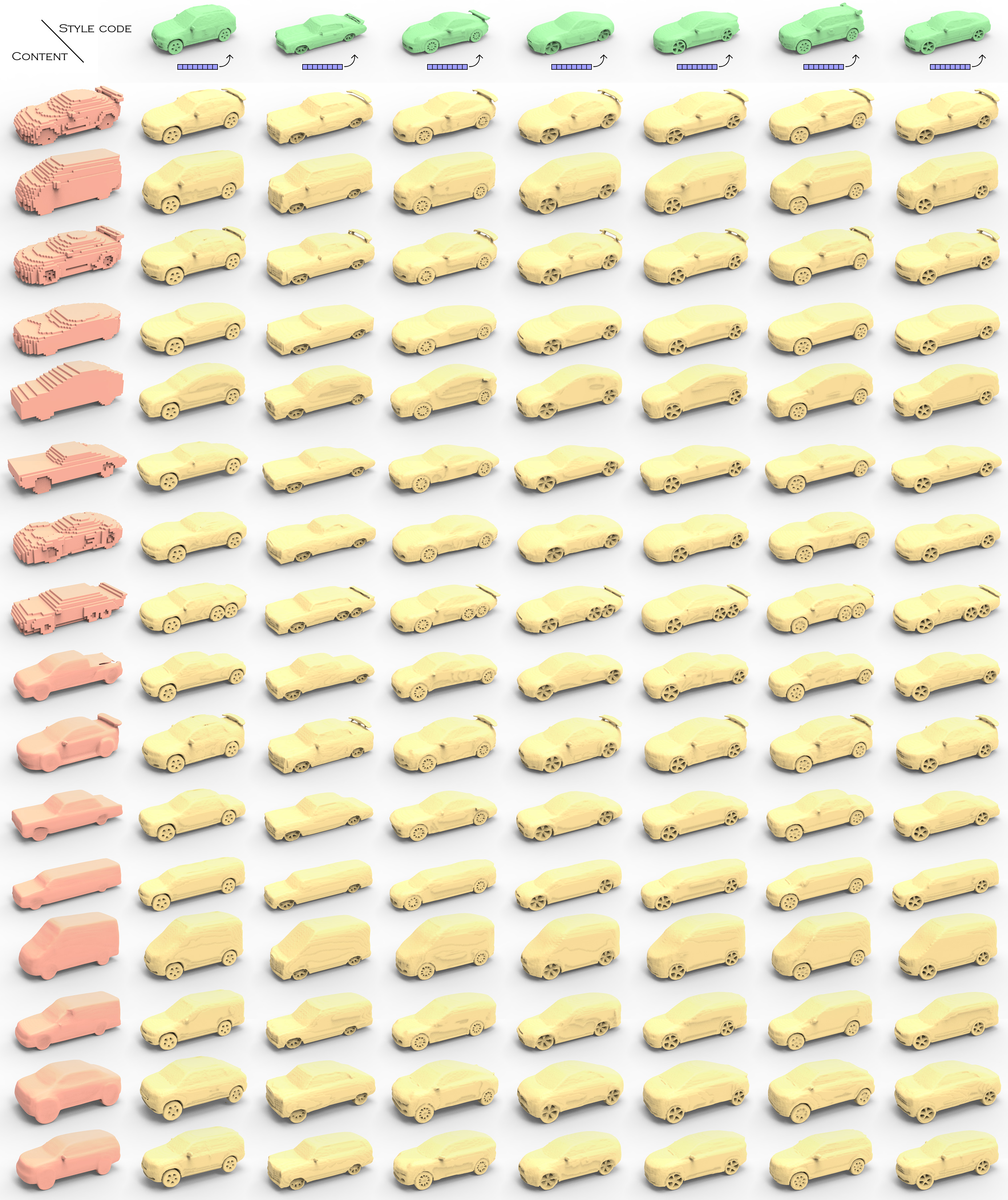}
\end{center}
\caption{
Results by upsampling coarse cars with different style codes. We show on the top the detailed shapes that correspond to the input style codes. The input shapes are coarse voxels in the first 8 rows, and downsampled versions of shapes generated by IM-GAN~\cite{IMNET} in the last 8 rows. The input resolution is $64^3$ and the output resolution is $256^3$.
}
\label{fig:supp_car}
\end{figure*}

%% file: figures/supp_plane.tex
\begin{figure*}[t!]
\begin{center}
\includegraphics[width=1.0\linewidth]{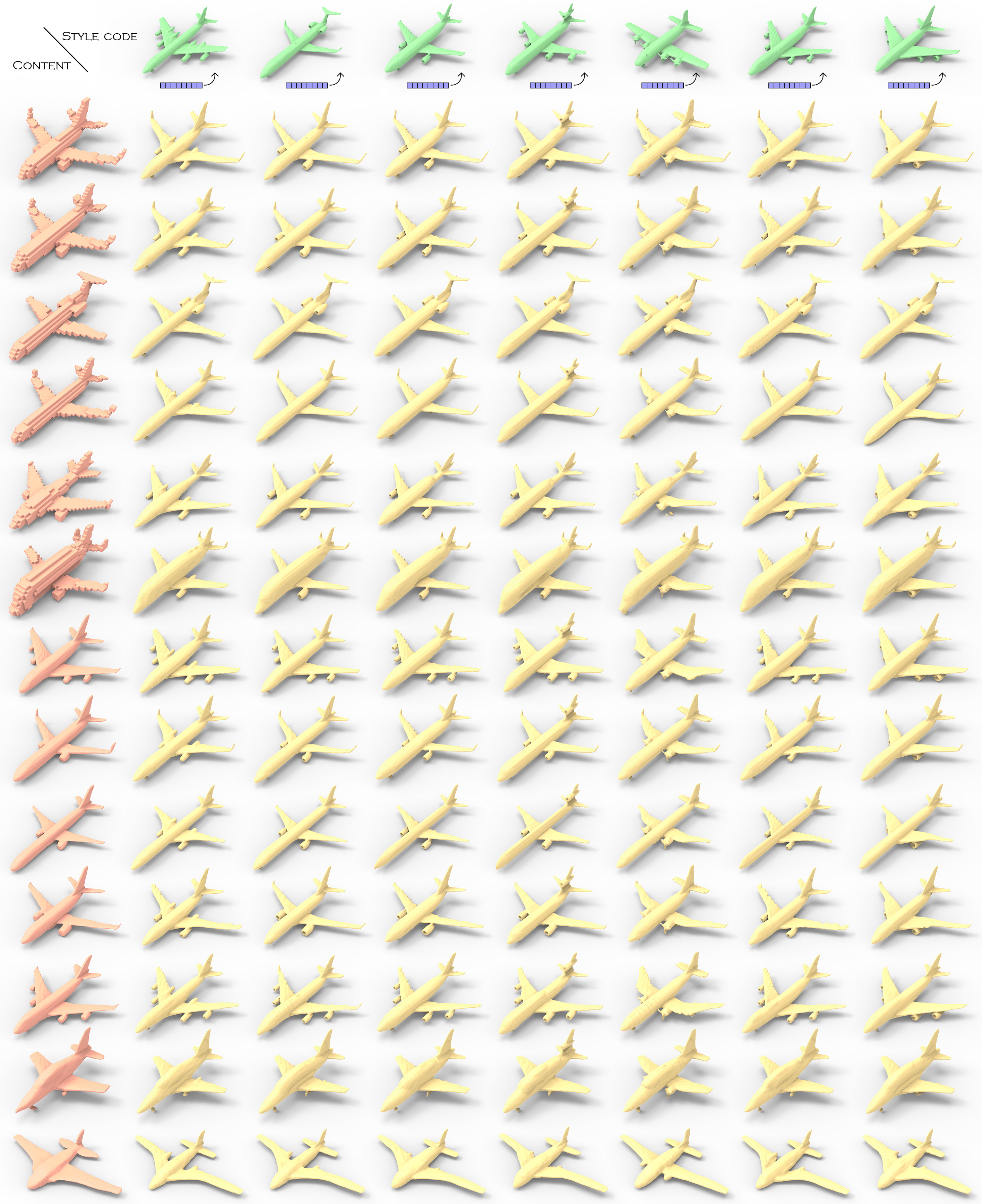}
\end{center}
\caption{
Results by upsampling coarse airplanes with different style codes. We show on the top the detailed shapes that correspond to the input style codes. The input shapes are coarse voxels in the first 6 rows, and downsampled versions of shapes generated by IM-GAN~\cite{IMNET} in the last 7 rows. The input resolution is $64^3$ and the output resolution is $256^3$.
}
\label{fig:supp_plane}
\end{figure*}

%% file: figures/supp_table.tex
\begin{figure*}[t!]
\begin{center}
\includegraphics[width=1.0\linewidth]{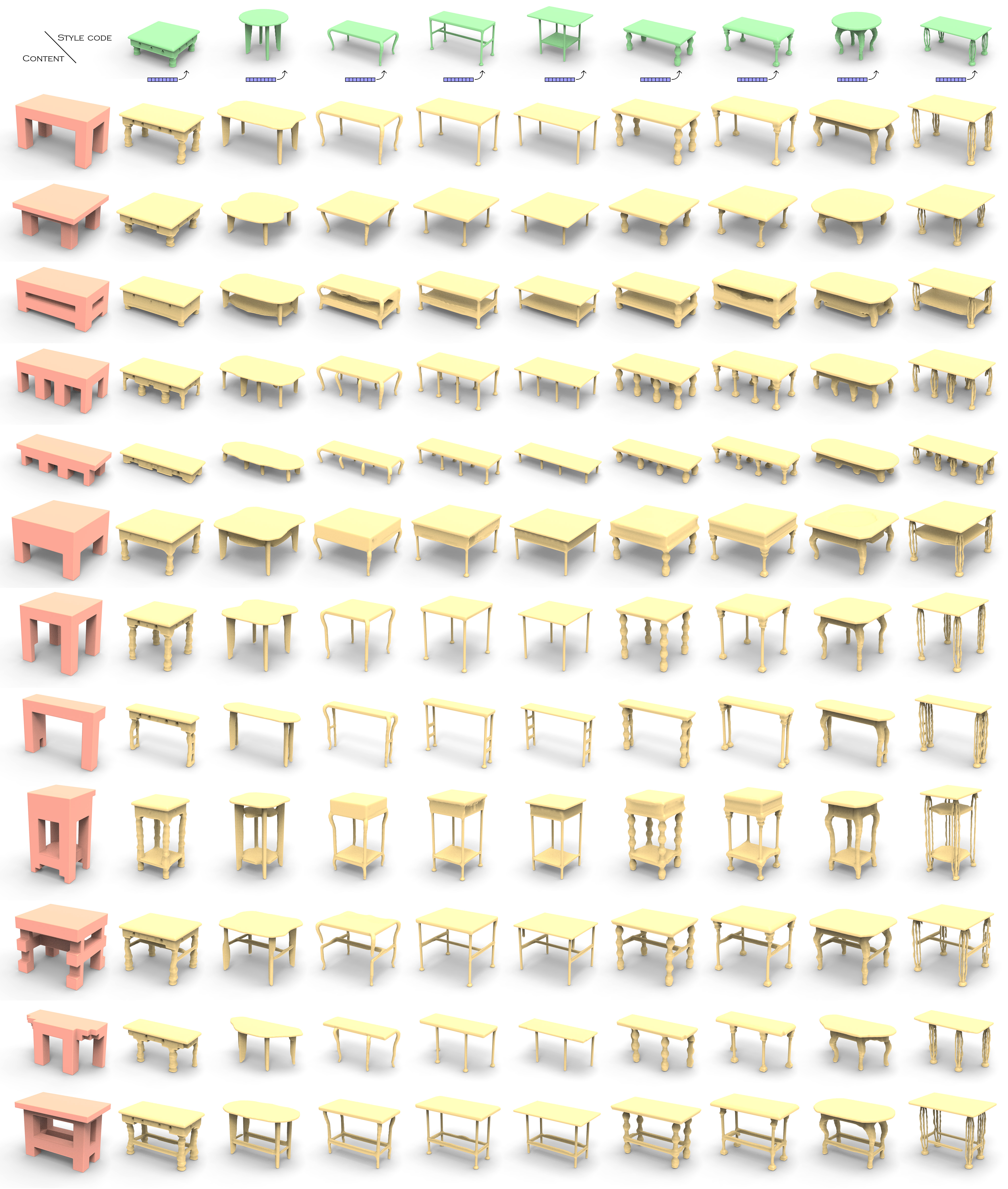}
\end{center}
\caption{
Results by upsampling coarse tables with different style codes. We show on the top the detailed shapes that correspond to the input style codes. The input shapes are coarse voxels. The input resolution is $16^3$ and the output resolution is $128^3$.
}
\label{fig:supp_table}
\end{figure*}

%% file: figures/supp_motor.tex
\begin{figure*}[t!]
\begin{center}
\includegraphics[width=1.0\linewidth]{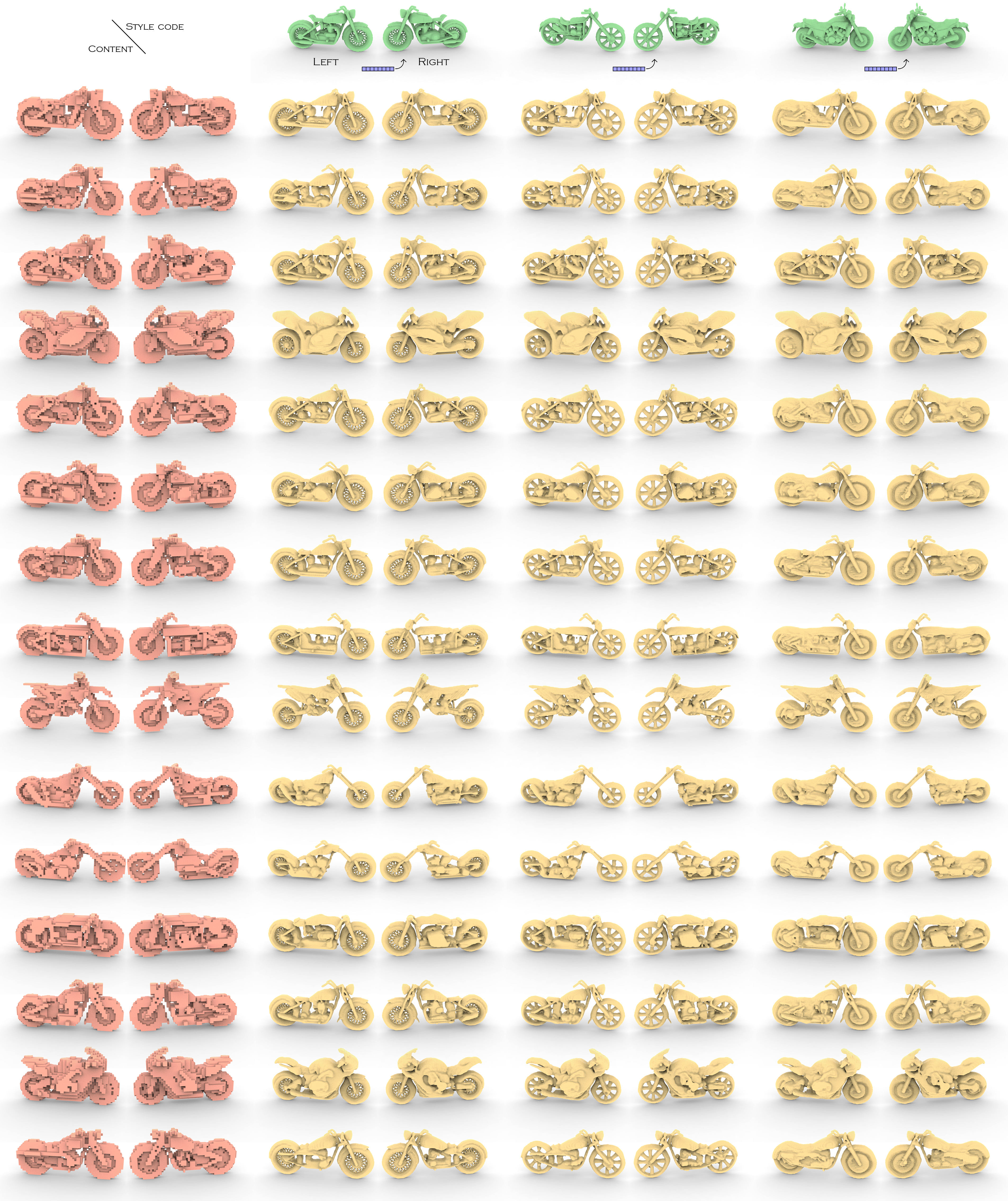}
\end{center}
\caption{
Results by upsampling coarse motorbikes with different style codes. Note that we lift the bilateral symmetry assumption for this category. We show on the top the detailed shapes that correspond to the input style codes. The input shapes are coarse voxels. The input resolution is $64^3$ and the output resolution is $256^3$.
}
\label{fig:supp_motor}
\end{figure*}

%% file: figures/supp_laptop.tex
\begin{figure*}[t!]
\begin{center}
\includegraphics[width=1.0\linewidth]{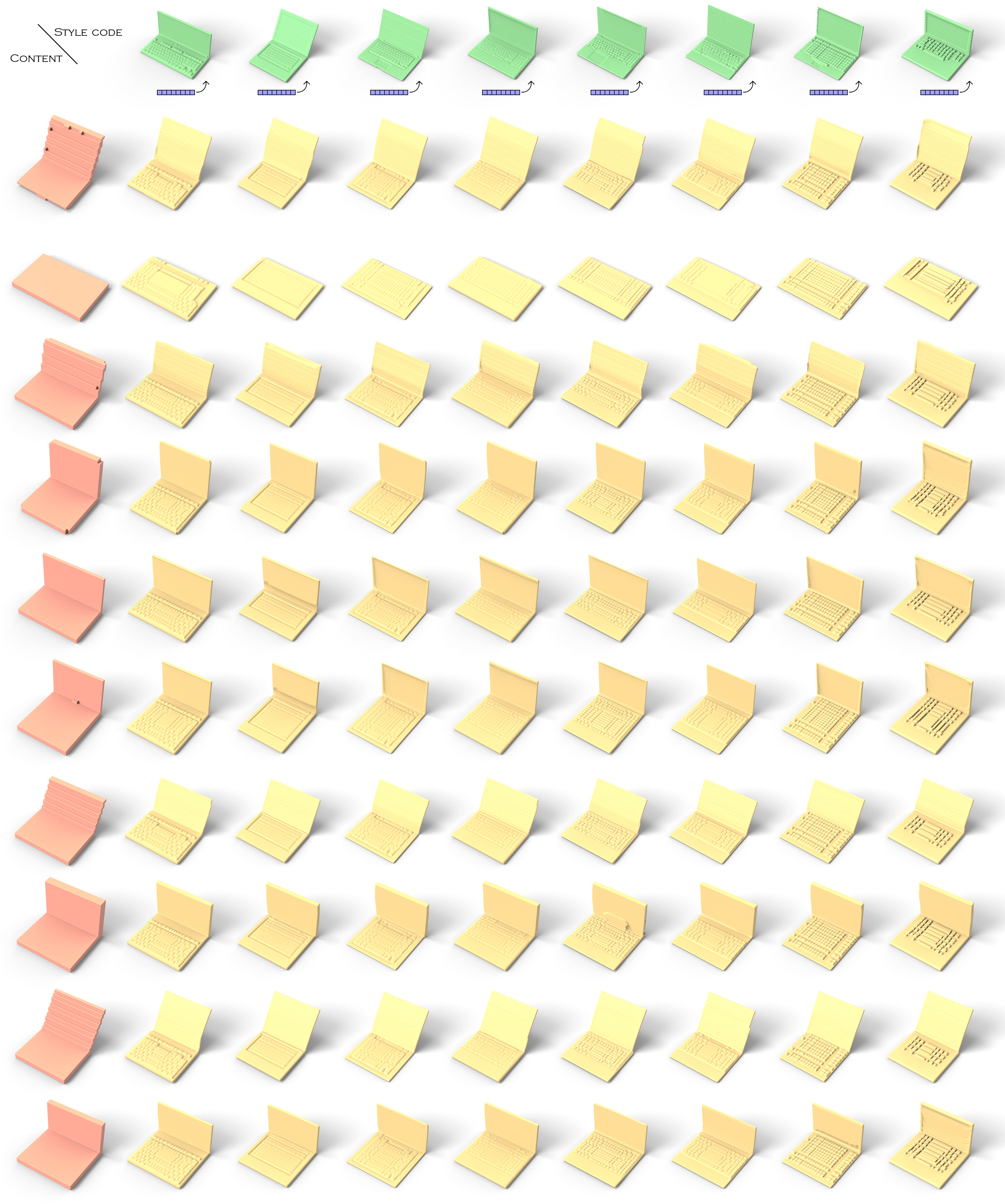}
\end{center}
\caption{
Results by upsampling coarse laptops with different style codes. Note that we lift the bilateral symmetry assumption for this category. We show on the top the detailed shapes that correspond to the input style codes. The input shapes are coarse voxels. The input resolution is $32^3$ and the output resolution is $256^3$.
}
\label{fig:supp_laptop}
\end{figure*}

%% file: figures/supp_plant.tex
\begin{figure*}[t!]
\begin{center}
\includegraphics[width=1.0\linewidth]{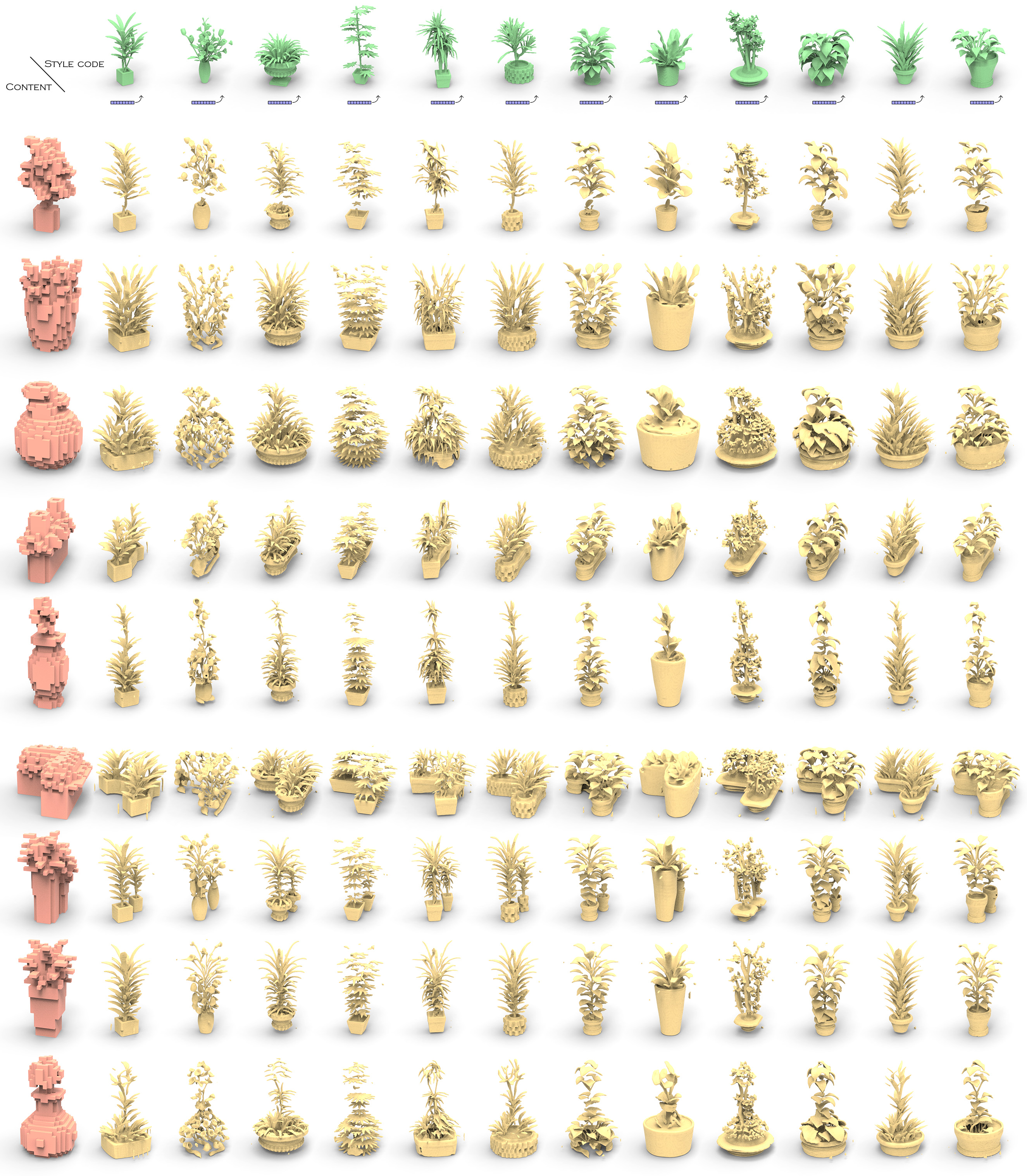}
\end{center}
\caption{
Results by upsampling coarse plants with different style codes. Note that we lift the bilateral symmetry assumption for this category. We show on the top the detailed shapes that correspond to the input style codes. The input shapes are coarse voxels. The input resolution is $32^3$ and the output resolution is $256^3$.
}
\label{fig:supp_plant}
\end{figure*}

%% file: figures/latent_plane.tex
\begin{figure*}[t!]
\begin{center}
\includegraphics[width=1.0\linewidth]{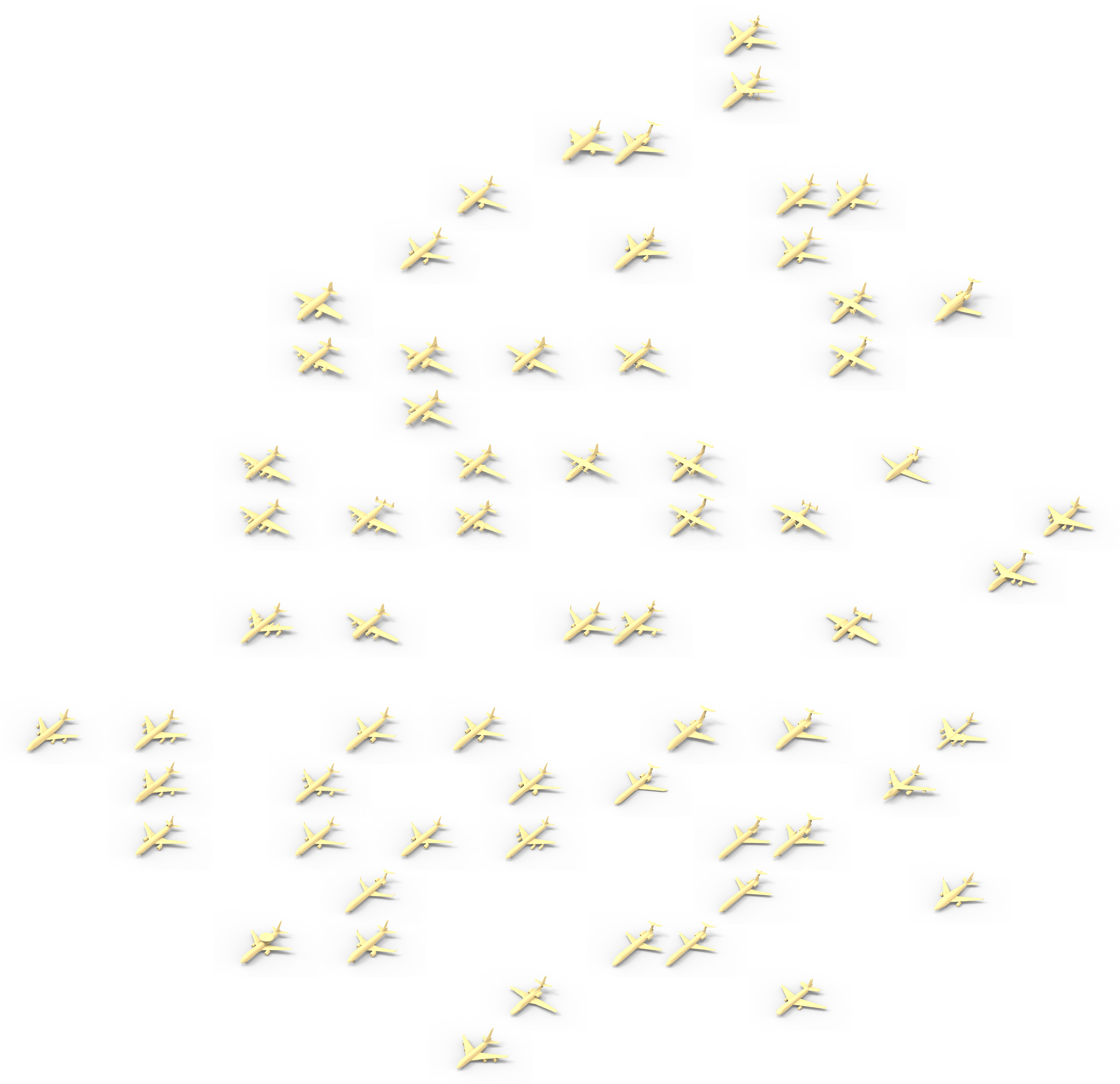}
\end{center}
\caption{
Visualization of 64 latent codes for airplanes via T-SNE embedding. For each latent code, the corresponding style shape is displayed in its location.
}
\label{fig:latent_plane}
\end{figure*}

%% file: figures/latent_car.tex
\begin{figure*}[t!]
\begin{center}
\includegraphics[width=1.0\linewidth]{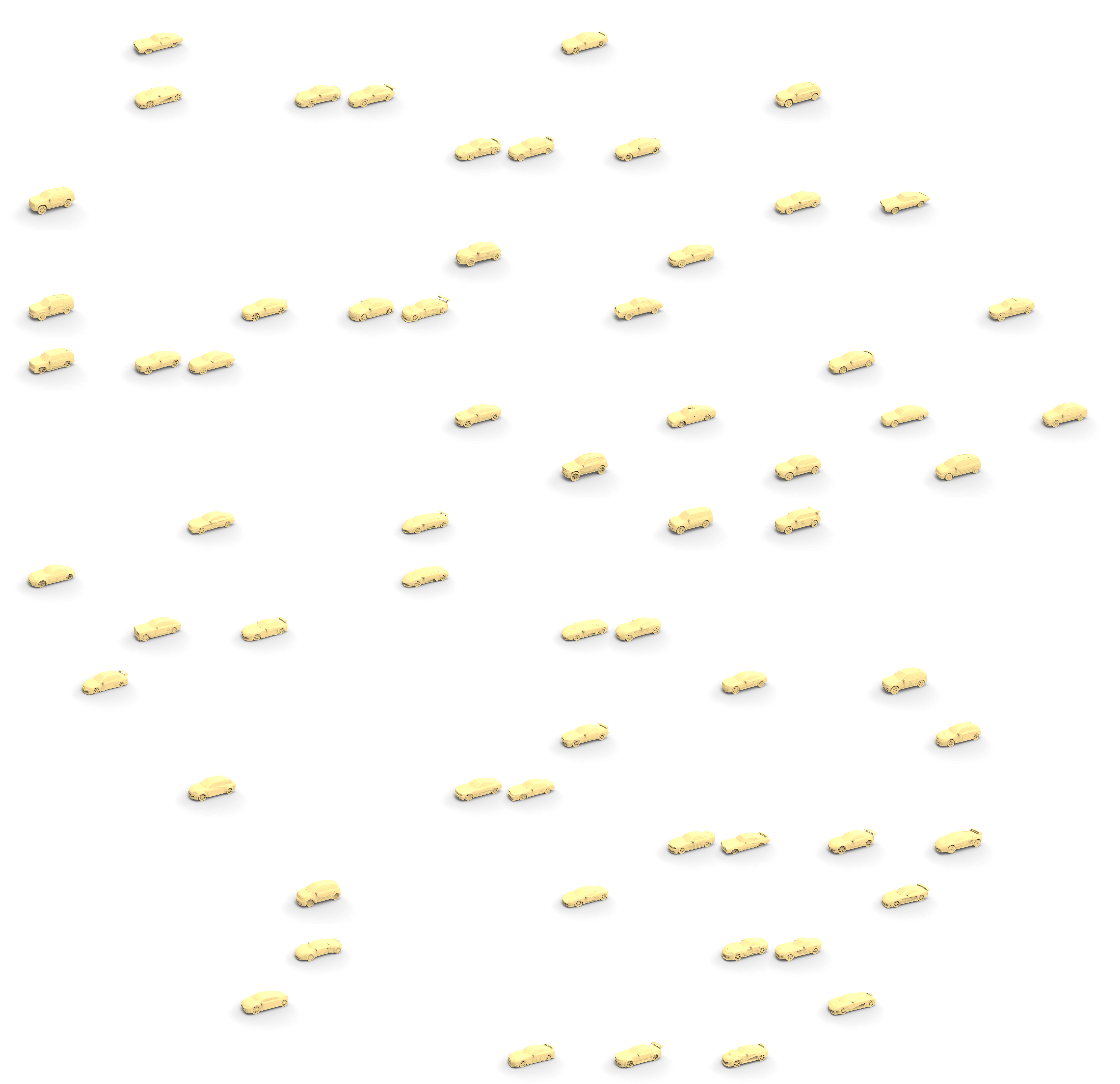}
\end{center}
\caption{
Visualization of 64 latent codes for cars via T-SNE embedding. For each latent code, the corresponding style shape is displayed in its location.
}
\label{fig:latent_car}
\end{figure*}

%% file: tables/numbers_chair.tex
\begin{table*}[!t]
\begin{center}
\resizebox{1.0\linewidth}{!}{
\begin{tabular}{l|c|c|c|c|c|c|c|c|c}
\hline
  & Strict-IOU $\uparrow$ & Loose-IOU $\uparrow$ & LP-IOU $\uparrow$ & LP-F-score $\uparrow$ & Div-IOU $\uparrow$ & Div-F-score $\uparrow$ & Cls-score $\downarrow$ & FID-all $\downarrow$ & FID-style $\downarrow$ \\
\hline
Recon. only								& 0.976 & 0.993 & 0.260 & 0.935 & 0.325 & 0.188 & 0.627 & 53.2 & 411.7 \\
No Gen. mask							& 0.655 & 0.792 & 0.452 & 0.973 & 0.825 & 0.806 & 0.672 & 121.9 & 379.9 \\
Strict Gen. mask						& 0.587 & 0.587 & 0.344 & 0.941 & 0.150 & 0.100 & 0.750 & 305.5 & 548.2 \\
No Dis. mask							& 0.145 & 0.167 &  N/A  &  N/A  &  N/A  &  N/A  & 0.843 & 2408.9 & 2714.1 \\
Conditional Dis. 1						& 0.947 & 0.981 & 0.259 & 0.949 & 0.291 & 0.194 & 0.593 & 51.3 & 402.7 \\
Conditional Dis. 3						& 0.928 & 0.977 & 0.246 & 0.963 & 0.197 & 0.206 & 0.603 & 55.8 & 418.2 \\
Proposed method*						& 0.673 & 0.805 & 0.432 & 0.973 & 0.800 & 0.816 & 0.644 & 113.1 & 372.5 \\
\hline
\hline
$\alpha = 0.0$, $N=16$					& 0.704 & 0.840 & 0.604 & 0.956 & 0.147 & 0.128 & 0.695 & 111.2 & 409.7 \\
$\alpha = 0.2$, $N=16$					& 0.583 & 0.750 & 0.527 & 0.971 & 0.875 & 0.934 & 0.667 & 115.5 & 371.5 \\
$\alpha = 0.5$, $N=16$					& 0.570 & 0.738 & 0.506 & 0.970 & 0.997 & 0.972 & 0.690 & 114.1 & 367.1 \\
No \small{$L_{GAN}^{global}$}, $N=16$	& 0.558 & 0.735 & 0.491 & 0.963 & 1.000 & 0.981 & 0.692 & 125.9 & 390.3 \\
\hline
\hline
$\alpha = 0.0$, $N=32$					& 0.763 & 0.864 & 0.551 & 0.962 & 0.184 & 0.156 & 0.598 & 131.2 & 391.7 \\
$\alpha = 0.2$, $N=32$					& 0.652 & 0.812 & 0.495 & 0.974 & 0.838 & 0.831 & 0.636 & 103.6 & 390.1 \\
$\alpha = 0.5$, $N=32$					& 0.598 & 0.757 & 0.470 & 0.974 & 0.934 & 0.934 & 0.662 & 111.1 & 380.0 \\
No \small{$L_{GAN}^{global}$}, $N=32$	& 0.561 & 0.728 & 0.462 & 0.969 & 0.997 & 0.984 & 0.690 & 109.1 & 368.2 \\
\hline
\hline
$\alpha = 0.0$, $N=64$					& 0.798 & 0.868 & 0.496 & 0.983 & 0.163 & 0.128 & 0.589 & 162.5 & 405.2 \\
$\alpha = 0.2$, $N=64$					& 0.781 & 0.864 & 0.423 & 0.985 & 0.353 & 0.334 & 0.619 & 109.2 & 370.3 \\
$\alpha = 0.5$, $N=64$*					& 0.673 & 0.805 & 0.432 & 0.973 & 0.800 & 0.816 & 0.644 & 113.1 & 372.5 \\
No \small{$L_{GAN}^{global}$}, $N=64$	& 0.578 & 0.741 & 0.426 & 0.965 & 0.950 & 0.988 & 0.669 & 116.3 & 381.8 \\
\hline
\hline
$\sigma = 0.0$							& 0.915 & 0.952 & 0.435 & 0.943 & 0.153 & 0.125 & 0.544 & 71.9 & 385.7 \\
$\sigma = 0.5$							& 0.869 & 0.919 & 0.493 & 0.952 & 0.172 & 0.144 & 0.580 & 101.2 & 379.5 \\
$\sigma = 1.0$*							& 0.673 & 0.805 & 0.432 & 0.973 & 0.800 & 0.816 & 0.644 & 113.1 & 372.5 \\
$\sigma = 1.5$							& 0.592 & 0.719 & 0.296 & 0.985 & 0.944 & 0.903 & 0.667 & 171.2 & 413.0 \\
$\sigma = 2.0$							& 0.565 & 0.614 & 0.208 & 0.982 & 0.575 & 0.666 & 0.711 & 244.8 & 482.7 \\
\hline
\hline
$\beta =  0.0$							& 0.730 & 0.815 & 0.279 & 0.967 & 0.178 & 0.269 & 0.669 & 129.9 & 391.1 \\
$\beta =  5.0$							& 0.652 & 0.785 & 0.448 & 0.974 & 0.822 & 0.775 & 0.642 & 135.4 & 378.7 \\
$\beta = 10.0$*							& 0.673 & 0.805 & 0.432 & 0.973 & 0.800 & 0.816 & 0.644 & 113.1 & 372.5 \\
$\beta = 15.0$							& 0.677 & 0.803 & 0.443 & 0.974 & 0.788 & 0.744 & 0.660 & 132.2 & 391.2 \\
$\beta = 20.0$							& 0.672 & 0.794 & 0.422 & 0.976 & 0.797 & 0.813 & 0.651 & 125.0 & 380.8 \\
\hline
\end{tabular}
}
\end{center}
\caption{Quantitative results for our ablation experiments on chairs. ``N/A'' is due to empty outputs. The models with * are the same model.}
\label{table:Ablation_chair}
\end{table*}

%% file: tables/numbers_car.tex
\begin{table*}[!t]
\begin{center}
\resizebox{1.0\linewidth}{!}{
\begin{tabular}{l|c|c|c|c|c|c|c|c|c}
\hline
  & Strict-IOU $\uparrow$ & Loose-IOU $\uparrow$ & LP-IOU $\uparrow$ & LP-F-score $\uparrow$ & Div-IOU $\uparrow$ & Div-F-score $\uparrow$ & Cls-score $\downarrow$ & FID-all $\downarrow$ & FID-style $\downarrow$ \\
\hline
Recon. only								& 0.991 & 0.998 & 0.760 & 0.998 & 0.172 & 0.084 & 0.493 & 153.4 & 457.0 \\
No Gen. mask							& 0.957 & 0.988 & 0.741 & 0.998 & 0.928 & 0.825 & 0.506 & 72.7 & 347.2 \\
Strict Gen. mask						& 0.829 & 0.829 & 0.751 & 0.995 & 0.159 & 0.084 & 0.538 & 303.2 & 569.3 \\
No Dis. mask							& 0.908 & 0.930 & 0.722 & 0.999 & 0.356 & 0.359 & 0.511 & 81.6 & 274.4 \\
Conditional Dis. 1						& 0.924 & 0.947 & 0.738 & 0.999 & 0.997 & 0.853 & 0.501 & 119.6 & 427.2 \\
Conditional Dis. 3						& 0.955 & 0.988 & 0.759 & 0.999 & 0.956 & 0.706 & 0.490 & 83.1 & 364.1 \\
Proposed method*						& 0.953 & 0.964 & 0.730 & 0.998 & 0.584 & 0.456 & 0.494 & 113.8 & 401.7 \\
\hline
\hline
$\alpha = 0.0$, $N=16$					& 0.882 & 0.987 & 0.832 & 0.996 & 0.275 & 0.238 & 0.600 & 1069.9 & 1478.7 \\
$\alpha = 0.2$, $N=16$					& 0.905 & 0.978 & 0.766 & 0.998 & 1.000 & 0.934 & 0.506 & 79.8 & 372.3 \\
$\alpha = 0.5$, $N=16$					& 0.909 & 0.975 & 0.772 & 0.999 & 1.000 & 0.941 & 0.492 & 84.5 & 377.7 \\
No \small{$L_{GAN}^{global}$}, $N=16$	& 0.900 & 0.972 & 0.764 & 0.998 & 1.000 & 0.947 & 0.500 & 79.6 & 377.2 \\
\hline
\hline
$\alpha = 0.0$, $N=32$					& 0.927 & 0.987 & 0.844 & 0.999 & 0.134 & 0.128 & 0.582 & 875.2 & 1251.7 \\
$\alpha = 0.2$, $N=32$					& 0.932 & 0.985 & 0.753 & 0.999 & 1.000 & 0.831 & 0.498 & 86.5 & 373.2 \\
$\alpha = 0.5$, $N=32$					& 0.922 & 0.979 & 0.756 & 0.999 & 1.000 & 0.909 & 0.507 & 77.2 & 356.1 \\
No \small{$L_{GAN}^{global}$}, $N=32$	& 0.910 & 0.970 & 0.745 & 0.998 & 1.000 & 0.928 & 0.497 & 68.0 & 357.1 \\
\hline
\hline
$\alpha = 0.0$, $N=64$					& 0.959 & 0.987 & 0.825 & 0.998 & 0.091 & 0.119 & 0.517 & 651.9 & 1019.5 \\
$\alpha = 0.2$, $N=64$*					& 0.955 & 0.988 & 0.759 & 0.999 & 0.956 & 0.706 & 0.490 & 83.1 & 364.1 \\
$\alpha = 0.5$, $N=64$					& 0.942 & 0.986 & 0.767 & 0.999 & 0.975 & 0.806 & 0.500 & 123.9 & 414.1 \\
No \small{$L_{GAN}^{global}$}, $N=64$	& 0.927 & 0.976 & 0.739 & 0.998 & 1.000 & 0.931 & 0.502 & 62.6 & 338.2 \\
\hline
\hline
$\sigma = 0.0$							& 0.977 & 0.994 & 0.763 & 0.995 & 0.119 & 0.075 & 0.499 & 223.3 & 548.7 \\
$\sigma = 0.5$							& 0.981 & 0.996 & 0.773 & 0.998 & 0.084 & 0.100 & 0.481 & 284.4 & 626.9 \\
$\sigma = 1.0$*							& 0.955 & 0.988 & 0.759 & 0.999 & 0.956 & 0.706 & 0.490 & 83.1 & 364.1 \\
$\sigma = 1.5$							& 0.938 & 0.983 & 0.750 & 0.999 & 0.991 & 0.838 & 0.490 & 85.6 & 363.9 \\
$\sigma = 2.0$							& 0.953 & 0.979 & 0.780 & 0.999 & 0.744 & 0.438 & 0.505 & 151.9 & 448.2 \\
\hline
\hline
$\beta =  0.0$							& 0.725 & 1.000 & 0.000 & 0.999 & 1.000 & 0.081 & 0.754 & 2759.8 & 3273.4 \\
$\beta =  5.0$							& 0.946 & 0.986 & 0.745 & 0.999 & 0.975 & 0.866 & 0.490 & 57.2 & 320.2 \\
$\beta = 10.0$*							& 0.955 & 0.988 & 0.759 & 0.999 & 0.956 & 0.706 & 0.490 & 83.1 & 364.1 \\
$\beta = 15.0$							& 0.958 & 0.989 & 0.753 & 0.999 & 0.894 & 0.753 & 0.500 & 75.8 & 350.0 \\
$\beta = 20.0$							& 0.950 & 0.985 & 0.750 & 0.998 & 0.994 & 0.878 & 0.505 & 64.7 & 334.8 \\
\hline
\end{tabular}
}
\end{center}
\caption{Quantitative results for our ablation experiments on cars. The models with * are the same model.}
\label{table:Ablation_car}
\end{table*}

%% file: tables/numbers_plane.tex
\begin{table*}[!t]
\begin{center}
\resizebox{1.0\linewidth}{!}{
\begin{tabular}{l|c|c|c|c|c|c|c|c|c}
\hline
  & Strict-IOU $\uparrow$ & Loose-IOU $\uparrow$ & LP-IOU $\uparrow$ & LP-F-score $\uparrow$ & Div-IOU $\uparrow$ & Div-F-score $\uparrow$ & Cls-score $\downarrow$ & FID-all $\downarrow$ & FID-style $\downarrow$ \\
\hline
Recon. only								& 0.966 & 0.980 & 0.465 & 0.999 & 0.166 & 0.100 & 0.493 & 64.8 & 328.6 \\
No Gen. mask							& 0.884 & 0.934 & 0.477 & 0.999 & 0.413 & 0.259 & 0.525 & 66.1 & 323.7 \\
Strict Gen. mask						& 0.487 & 0.487 & 0.380 & 0.974 & 0.069 & 0.072 & 0.642 & 1252.5 & 1196.9 \\
No Dis. mask							& 0.508 & 0.564 & 0.277 & 0.998 & 0.084 & 0.141 & 0.539 & 552.6 & 859.4 \\
Conditional Dis. 1						& 0.782 & 0.855 & 0.477 & 0.997 & 0.347 & 0.294 & 0.493 & 667.5 & 773.1 \\
Conditional Dis. 3						& 0.809 & 0.854 & 0.443 & 0.996 & 0.094 & 0.119 & 0.524 & 717.4 & 795.9 \\
Proposed method*						& 0.875 & 0.947 & 0.474 & 0.998 & 0.516 & 0.353 & 0.487 & 57.3 & 340.9 \\
\hline
\hline
$\alpha = 0.0$, $N=16$					& 0.843 & 0.921 & 0.510 & 0.999 & 0.069 & 0.059 & 0.516 & 93.9 & 331.7 \\
$\alpha = 0.1$, $N=16$					& 0.764 & 0.890 & 0.487 & 0.997 & 0.825 & 0.659 & 0.502 & 100.6 & 307.7 \\
$\alpha = 0.2$, $N=16$					& 0.720 & 0.845 & 0.501 & 0.990 & 0.994 & 0.897 & 0.504 & 106.3 & 308.0 \\
No \small{$L_{GAN}^{global}$}, $N=16$	& 0.657 & 0.805 & 0.516 & 0.986 & 1.000 & 0.947 & 0.504 & 132.5 & 354.9 \\
\hline
\hline
$\alpha = 0.0$, $N=32$					& 0.883 & 0.946 & 0.503 & 1.000 & 0.059 & 0.066 & 0.515 & 95.2 & 350.9 \\
$\alpha = 0.1$, $N=32$					& 0.835 & 0.926 & 0.459 & 0.999 & 0.734 & 0.538 & 0.503 & 71.4 & 329.4 \\
$\alpha = 0.2$, $N=32$					& 0.777 & 0.887 & 0.481 & 0.997 & 0.947 & 0.788 & 0.520 & 79.5 & 325.6 \\
No \small{$L_{GAN}^{global}$}, $N=32$	& 0.675 & 0.818 & 0.493 & 0.989 & 1.000 & 0.941 & 0.493 & 135.2 & 377.0 \\
\hline
\hline
$\alpha = 0.0$, $N=64$					& 0.898 & 0.959 & 0.499 & 0.999 & 0.059 & 0.056 & 0.503 & 80.2 & 353.6 \\
$\alpha = 0.1$, $N=64$*					& 0.875 & 0.947 & 0.474 & 0.998 & 0.516 & 0.353 & 0.487 & 57.3 & 340.9 \\
$\alpha = 0.2$, $N=64$					& 0.831 & 0.921 & 0.463 & 0.998 & 0.756 & 0.600 & 0.498 & 67.7 & 332.0 \\
No \small{$L_{GAN}^{global}$}, $N=64$	& 0.707 & 0.833 & 0.478 & 0.991 & 0.997 & 0.934 & 0.489 & 105.8 & 354.8 \\
\hline
\hline
$\sigma = 0.0$							& 0.802 & 0.892 & 0.422 & 0.988 & 0.059 & 0.041 & 0.495 & 205.1 & 363.0 \\
$\sigma = 0.5$							& 0.883 & 0.911 & 0.464 & 0.997 & 0.066 & 0.084 & 0.508 & 81.7 & 327.2 \\
$\sigma = 1.0$*							& 0.875 & 0.947 & 0.474 & 0.998 & 0.516 & 0.353 & 0.487 & 57.3 & 340.9 \\
$\sigma = 1.5$							& 0.845 & 0.899 & 0.451 & 0.997 & 0.469 & 0.409 & 0.518 & 175.2 & 372.7 \\
$\sigma = 2.0$							& 0.730 & 0.767 & 0.534 & 0.993 & 0.278 & 0.384 & 0.549 & 847.1 & 818.3 \\
\hline
\hline
$\beta =  0.0$							& 0.384 & 1.000 & 0.000 & 0.940 & 0.659 & 0.050 & 0.811 & 6342.0 & 5491.9 \\
$\beta =  5.0$							& 0.892 & 0.948 & 0.460 & 0.999 & 0.325 & 0.188 & 0.492 & 73.2 & 310.0 \\
$\beta = 10.0$*							& 0.875 & 0.947 & 0.474 & 0.998 & 0.516 & 0.353 & 0.487 & 57.3 & 340.9 \\
$\beta = 15.0$							& 0.895 & 0.952 & 0.468 & 0.999 & 0.450 & 0.319 & 0.500 & 63.8 & 354.6 \\
$\beta = 20.0$							& 0.872 & 0.946 & 0.459 & 0.998 & 0.531 & 0.475 & 0.517 & 69.2 & 311.6 \\
\hline
\end{tabular}
}
\end{center}
\caption{Quantitative results for our ablation experiments on airplanes. The models with * are the same model.}
\label{table:Ablation_plane}
\end{table*}